\documentclass[11pt]{article}

\usepackage[english]{babel}
\usepackage{bbm}
\usepackage{amsmath, amssymb, amsthm}
\usepackage{mathrsfs}
\usepackage{comment}
\usepackage{graphicx} %% Package for Figure
\usepackage{subcaption}
\usepackage{float}
\usepackage{xcolor}
\usepackage{bm}
\usepackage[letterpaper,top=2cm,bottom=2cm,left=2.7cm,right=2.7cm,marginparwidth=1.75cm]{geometry}
\usepackage{multirow}

\newcommand{\norm}[1]{\left\lVert#1\right\rVert}
\usepackage{graphicx}
\usepackage[colorlinks=true, allcolors=blue]{hyperref}
\usepackage{algorithmic}
\usepackage{indentfirst}  
\usepackage{booktabs}
\usepackage{array}

\usepackage{amsmath,amssymb,amsthm}
\usepackage{algorithm}

\theoremstyle{remark}
\newtheorem{remark}{Remark}

\newtheorem{theorem}{Theorem}[section]
\usepackage{authblk}

\newcommand{\E}{\mathbb{E}}
\newcommand{\Sop}{\mathcal{S}}
\newcommand{\Gop}{\mathcal{G}}

\newcommand{\MSE}{\operatorname{MSE}}
\newcommand{\PI}{\mathrm{PI}}

\usepackage{float}
\usepackage{caption}

\title{\Large{StablePDENet: Enhancing Neural Operator Stability through Physics-Informed Residual-Sensitivity Regularization}}%A Physics-Informed Neural Operator Stable under Perturbation
\author[1]{Chutian Huang}
\author[1]{Chang Ma\thanks{Corresponding authors: changma@ust.hk (Chang Ma), maxiang@ust.hk (Yang Xiang)}}
\author[1]{Kaibo Wang}
\author[1,2]{Yang Xiang$^{*}$}
\affil[1]{Department of Mathematics, The Hong Kong University of Science and Technology, Clear Water Bay, Kowloon,
	Hong Kong}
\affil[2]{Algorithms of Machine Learning and Autonomous Driving Research Lab, HKUST Shenzhen-Hong Kong
	Collaborative Innovation Research Institute, Shenzhen, China}

\date{}
\begin{document}
\graphicspath{{figures/}}
\maketitle

\begin{abstract}
Learning solution operators for differential equations with neural networks has shown great potential in scientific computing, but ensuring their stability under input perturbations remains a critical challenge. 
We introduce the StablePDENet, a physics-informed
adversarial training method that regularizes the residual sensitivity
with respect to an input perturbation.
The operator learning task is formulated as a min--max optimization problem, where the inner model searches admissible input perturbations by physics-based projected-gradient adversary,
while the outer problem combines
the attacked physics loss with a normalized residual-sensitivity penalty.
Moreover, residual-sensitivity regularization is included to ensure that the local Lipschitz constant of the learned operator is  a more accurate 
approximation to that of the exact operator.
We evaluate the StablePDENet on several benchmark problems. 
Compared with
PI-DeepONet and its adversarially trained variant, StablePDENet achieves higher
accuracy under adversarial input perturbations while maintaining competitive
accuracy on clean inputs. 
The numerical results also demonstrate that the StablePDENet can effectively improve the generalization accuracy for operator learning. The Helmholtz study further distinguishes learned-model sensitivity from amplification intrinsic to an ill-conditioned solution operator. The results support
residual-sensitivity regularization as a practical route to more stable and physically consistent neural PDE operators. 
\end{abstract}


\section{Introduction}

In recent years, neural network-based approaches have shown promising performance in addressing the forward and inverse problems of partial differential equations (PDEs) raised in science and engineering, including fluid dynamics, climate modeling, molecular dynamics and materials mechanics \cite{brunton2020machine, zhang2018deep, bao2025pfwnn, kashinath2021physics, Chen:20}.  PINNs and related
variational methods incorporate governing equations into the training
objective, while DeepONet and the Fourier neural operator learn mappings
between function spaces across families of PDEs
\cite{RAISSI2019686,yu2018deep,zang2020weak,cite-key,li2021fourier}. Despite
this maturity, the input stability of learned solution operators
under perturbations to physical inputs remains insufficiently understood.

This paper studies input stability, namely the Lipschitz stability of
a learned operator with respect to perturbations in its physical inputs. Let
$\Sop:\mathcal X\to\mathcal Y$ denote the physical solution operator and let
$\Gop_\theta$ be its learned approximation. On an admissible neighborhood
$\mathcal D\subset\mathcal X$, the learned operator is Lipschitz stable if
\begin{equation*}
	\left\|\Gop_\theta(a_1)-\Gop_\theta(a_2)\right\|_{\mathcal Y}
	\leq
	L_\theta
	\left\|a_1-a_2\right\|_{\mathcal X},
	\qquad a_1,a_2\in\mathcal D.
\end{equation*}
Here, $L_\theta$ is a Lipschitz constant of the learned model and quantifies the amplification of input perturbations. Controlling such amplification is important because even minor input perturbations can lead to substantial prediction errors, spurious oscillations, or unstable numerical behavior \cite{higham2002accuracy}. However, the physical solution operator $\Sop$ may itself be
highly sensitive. Therefore, our objective is to make the input response of $\Gop_\theta$ more
consistent with that of $\Sop$ instead of minimizing $L_\theta$
indiscriminately.

Classical notions of numerical stability depend on both the PDE and its
discretization. The Lax equivalence theorem applies to consistent
discretizations of well-posed linear problems, while other numerical
formulations require different, problem-specific stability criteria \cite{lax2005survey,iserles2009first,morton2005numerical}. These classical
notions motivate the control of perturbation amplification, but this work does
not claim an equivalent mesh-uniform convergence theorem or numerical
stability certificate for a trained neural operator.

Unlike traditional numerical solvers, the stability properties of neural network–based PDE solvers remain insufficiently understood and lack a systematic theoretical framework despite the empirical success in many cases.
Few works explicitly address stability and incorporate stability constraints into their training objectives or architectures, thereby leaving no guarantee that the learned solution remains stable under perturbations. 
Recently,  uncertainty quantification \cite{PSAROS2023111902} has emerged as a practical tool to assess the reliability of deep learning PDE solvers. It helps identify regions where neural network predictions may be unstable or unreliable. Nevertheless, it serves as a diagnostic tool but does not enforce stability directly. 

The stability of neural networks has emerged as a critical safety concern in machine learning, particularly when deployed in high-stakes applications. Key studies focus on security-critical fields such as autonomous driving, facial recognition, and natural language processing \cite{ morris2020textattack,deng2020analysis,sharif2016accessorize}. 
The fragility of neural networks was demonstrated by Goodfellow et al. \cite{Goodfellow2014ExplainingAH}, who showed that imperceptible perturbations --- constructed by leveraging gradient information --- could deceive classifiers into mislabeling images with high confidence, such as identifying a panda as a gibbon.
To mitigate these risks, adversarial training \cite{madry2018towards,wang2024diffhammer} and other defense mechanisms \cite{papernot2016distillation,chivukula2020game} have been proposed to improve the robustness of models. 
Adversarial learning provides an offensive yet rigorous framework by explicitly constructing worst-case perturbations that maximize model error. 
Recent advances in adversarial learning for PDEs have further emphasized the
need to understand how learned solution models respond to variations in
physical inputs. 
Gradient-based adversarial optimization, one of the most widely studied methods for generating adversarial examples, provides a systematic means of probing sensitive input directions \cite{Goodfellow2014ExplainingAH,madry2018towards}. The
interpretation differs from classification: perturbing a PDE input changes the underlying
physical problem and generally changes its exact solution. Stability therefore
concerns physically consistent solution sensitivity rather than label
invariance.
Preliminary results on this gradient-based mechanism were reported in the doctoral thesis of Huang \cite{huang2025thesis}, which formulated operator learning as a min--max optimization problem. In this framework, worst-case input perturbations are generated along the physics-residual gradient toward regions of large residuals, thereby enabling the model to maintain consistent performance on both original and adversarially perturbed inputs.
Another
related gradient-based mechanism appears in RAMS
\cite{ouyang2026rams}, which moves sampled
input functions by attack learning. It is designed to facilitate adaptive sampling and improve training efficiency, rather than to enhance the stability of the learned solution operator.

In this work, we propose the StablePDENet, a novel framework for studying and
improving the input stability of neural operators for differential equations.
The StablePDENet uses adversarially perturbed physical inputs to probe operator
sensitivity and explicitly regularizes the induced residual sensitivity within a
min--max operator-learning formulation. The
inner problem searches for admissible physical-input perturbations, and the
outer problem combines the attacked physics loss with a normalized
residual-sensitivity regularizer. This construction controls spurious learned
responses without suppressing sensitivity intrinsic to the PDE. 
Under
explicit inverse-stability assumptions, the analysis provides conditional
residual-to-error estimates. Moreover, a residual-sensitivity regularizer is included to ensure that the local Lipschitz constant of the learned neural operator is  a more accurate 
	 approximation to that of the exact operator. Numerically, the StablePDENet improves perturbed
accuracy while retaining competitive clean-input accuracy on some benchmarks, and it yields a more
faithful empirical estimate of  local Lipschitz constant that more closely matches that of the physical solution operator.
The results further indicate that the StablePDENet can effectively improve the generalization accuracy for operator learning.

The remainder of the paper is organized as follows. We start with introducing our notation and preliminary results in Section \ref{sec:formulation}. In Section \ref{sec:stablepdenet}, the StablePDENet framework and algorithm are proposed for enhancing the stability of the operator learning methods, which are based on the adversarial training with residual-sensitivity regularizer. In Section \ref{section:03}, the accuracy and efficiency of the proposed method are examined through several numerical experiments. Finally, this paper is concluded with general discussions in Section \ref{section:04}.

\section{Problem Formulation}
\label{sec:formulation}
To conduct our research on the stability of neural operator learning, we first introduce the methods employed in this paper. The neural operator framework used here serves as an example and can be replaced with any operator learning framework.
\subsection{Operator learning for parametric PDEs}

Let $a\in\mathcal{X}$ denote the input of the PDE solution operator. Depending on
the benchmark, $a$ may be a spatially varying source, an initial condition, or a
finite-dimensional coefficient vector. Here, $\mathcal{X}$ is the admissible
input space, $\mathcal{Y}$ is the solution space, and
$u=\Sop(a)\in\mathcal{Y}$ denotes the corresponding exact or high-fidelity
solution. The solution operator is defined implicitly by
\begin{equation}
	\mathcal{F}(u,a)=0,
	\qquad
	\Sop:\mathcal{X}\to\mathcal{Y},
	\quad a\mapsto u=\Sop(a).
	\label{eq:abstract-pde}
\end{equation}

To include the differential equation and its auxiliary conditions in one
notation, let $\mathcal{B}$ be the finite index set of residual blocks enforced
for the equation under consideration. Typical elements of $\mathcal{B}$ are
$\mathrm{PDE}$ for the interior differential equation, $\mathrm{BC}$ for a
boundary condition, and $\mathrm{IC}$ for an initial condition, i.e., $\mathcal{B}=\{\mathrm{PDE},\mathrm{BC},\mathrm{IC}\}$. For each $q\in\mathcal{B}$, let
$\mathcal{F}_q:\mathcal{Y}\times\mathcal{X}\to\mathcal{Z}_q$ be the residual
operator associated with block $q$, where $\mathcal{Z}_q$ is its residual
space. We then write
\begin{equation}
	\mathcal{F}(u,a)
	:=\bigl(\mathcal{F}_q(u,a)\bigr)_{q\in\mathcal{B}}
	\in\prod_{q\in\mathcal{B}}\mathcal{Z}_q,
	\qquad
	\mathcal{F}_q(u,a)=0
	\quad\text{for every }q\in\mathcal{B}.
	\label{eq:residual-blocks}
\end{equation}
For brevity, write $Z:= \prod_{q\in\mathcal{B}}\mathcal{Z}_q$.

When boundary or initial conditions are imposed softly,
$\mathcal{F}$ denotes the augmented residual operator containing the
interior and auxiliary-condition blocks. Any inverse-stability assumption
below therefore concerns this augmented operator or its linearization. Such
inverse stability is an explicit analytical assumption and is not implied
merely by obtaining a small empirical training loss.

Let $m$ be the branch-input dimension and let
$x_1^{\rm br},\ldots,x_m^{\rm br}$ denote the corresponding branch-sensor
locations. The discrete branch input is
\begin{equation}
	\mathbf{a}_m=(a_1,\ldots,a_m),
	\qquad
	a_i=a(x_i^{\rm br}),
\end{equation}
where $a_i$
denotes its $i$th component. 
An exemplary framework in operator learning is first presented for context.
We use a self-supervised backend network, the Physics-Informed DeepONet (PI-DeepONet) \cite{wang2021learning}. 
Following DeepONet, PI-DeepONet consists of two main components: a branch network and a trunk network. The branch network takes sampled values of an input function, such as an initial condition, boundary condition, coefficient, or source term, and encodes them into coefficients. The trunk network takes the output evaluation coordinate and returns basis functions evaluated at that coordinate. 
For a scalar-output PI-DeepONet, the
prediction $\mathcal{G}_\theta$ at an output coordinate $\xi\in\Omega_{\rm out}$ is
\begin{equation}
	\widehat{u}_\theta(a)(\xi)
	=\Gop_\theta(a)(\xi)
	=\sum_{j=1}^{p}
	b_{\theta,j}(\mathbf{a}_m)\,
	t_{\theta,j}(\xi)+b_0.
	\label{eq:deeponet}
\end{equation}
Here, $\Omega_{\rm out}$ is the domain on which the solution is queried,
$b_\theta:\mathbb{R}^{m}\to\mathbb{R}^{p}$ is the branch network, and
$t_\theta:\Omega_{\rm out}\to\mathbb{R}^{p}$ is the trunk network.
The quantities $b_{\theta,j}$ and $t_{\theta,j}$ are their $j$th scalar
components, $p$ is the shared latent dimension, $b_0\in\mathbb{R}$ is a
trainable output bias, and $\theta$ collects all trainable parameters. The
coordinate $\xi$ is spatial for a steady problem and space--time for an
evolutionary problem.

For each block $q\in\mathcal{B}$, let
$\mathcal{C}_q=\{\eta_{q,\ell}\}_{\ell=1}^{N_q}$ be its collocation set, with
$N_q=|\mathcal{C}_q|$, where
$\eta_{q,\ell}$ lies in the appropriate interior, boundary, initial-time, or
periodic domain. The pointwise residual of the neural prediction is
\begin{equation}
	r_{\theta,q}(a;\eta_{q,\ell})
	:=
	\left[
	\mathcal{F}_q\bigl(\Gop_\theta(a),a\bigr)
	\right](\eta_{q,\ell}),
	\qquad
	q\in\mathcal{B}.
	\label{eq:block-residual}
\end{equation}
For each residual block, collect the sampled residuals in the vector
\begin{equation}
		\mathbf r_{\theta,q}(a)
		:=
		\left(
		r_{\theta,q}(a;\eta_{q,\ell})
		\right)_{\ell=1}^{N_q}.
\end{equation}
Its normalized discrete $\ell_2$ norm is
\begin{equation}
		\left\|\mathbf r_{\theta,q}(a)\right\|_{2,N_q}^{2}
		:=
		\frac{1}{N_q}
		\sum_{\ell=1}^{N_q}
		\left\|
		r_{\theta,q}(a;\eta_{q,\ell})
		\right\|_2^2.
	\label{eq:block-discrete-l2}
\end{equation}
Here, $\|\cdot\|_2$ is the Euclidean norm over the residual channels; for
	a scalar residual, it reduces to the ordinary square. The weighted physics
	loss is therefore
\begin{equation}
		\mathcal{L}_{\rm phys}(\theta;a)
		=
		\sum_{q\in\mathcal{B}}
		w_q
		\left\|\mathbf r_{\theta,q}(a)\right\|_{2,N_q}^{2},
		\qquad w_q>0.
	\label{eq:physics-loss}
\end{equation}
where $w_q$ is the prescribed equation-specific weight of block $q$. Branch
sensors, residual collocation sets, and reference-solver grids have distinct
roles even when an experiment places some of them on the same numerical grid.
No paired solution labels $(a,\Sop(a))$ are used during training; instead, the
PDE and auxiliary residuals are driven toward zero.

In order to clarify the notation, let us take the one-dimensional Poisson equation as an example,
\begin{equation}
	-u_{xx}(x)=f(x),\qquad x\in[0,1],\qquad u(0)=u(1)=0.
\end{equation}
We learn the operator $\mathcal{G}_\theta:f\mapsto u$. Let
$\mathbf f^{(i)}=[f^{(i)}(x_1),\ldots,f^{(i)}(x_m)]^\top$ be the sampled branch input for the $i$-th source function, and let $\{\xi_\ell\}_{\ell=1}^{n_c}$ be the collocation points for the PDE residual. The loss is
\begin{equation}\label{equ:possion_1d}
	\begin{aligned}
		\mathcal{L}_{\rm phys}(\theta)
		&=\mathcal{L}_{\mathrm{PDE}}(\theta)+\mathcal{L}_{\mathrm{BC}}(\theta)\\
		&=\frac{1}{Nn_c}\sum_{i=1}^{N}\sum_{\ell=1}^{n_c}
		\left|
		-\frac{\partial^2}{\partial \xi^2}\mathcal{G}_\theta\!\left(\mathbf f^{(i)}\right)(\xi_\ell)
		-f^{(i)}(\xi_\ell)
		\right|^2\\
		&\quad+
		\frac{1}{N}\sum_{i=1}^{N}\left(
		\left|\mathcal{G}_\theta\!\left(\mathbf f^{(i)}\right)(0)\right|^2+
		\left|\mathcal{G}_\theta\!\left(\mathbf f^{(i)}\right)(1)\right|^2
		\right).
	\end{aligned}
\end{equation}
The branch sensors $\{x_j\}_{j=1}^{m}$ and residual collocation points $\{\xi_\ell\}_{\ell=1}^{n_c}$ are conceptually distinct. In some experiments they are chosen from the same grid for convenience; in general they need not coincide.

\section{StablePDENet}
\label{sec:stablepdenet}
\begin{figure}[H]
	\centering
	\includegraphics[width=\linewidth]{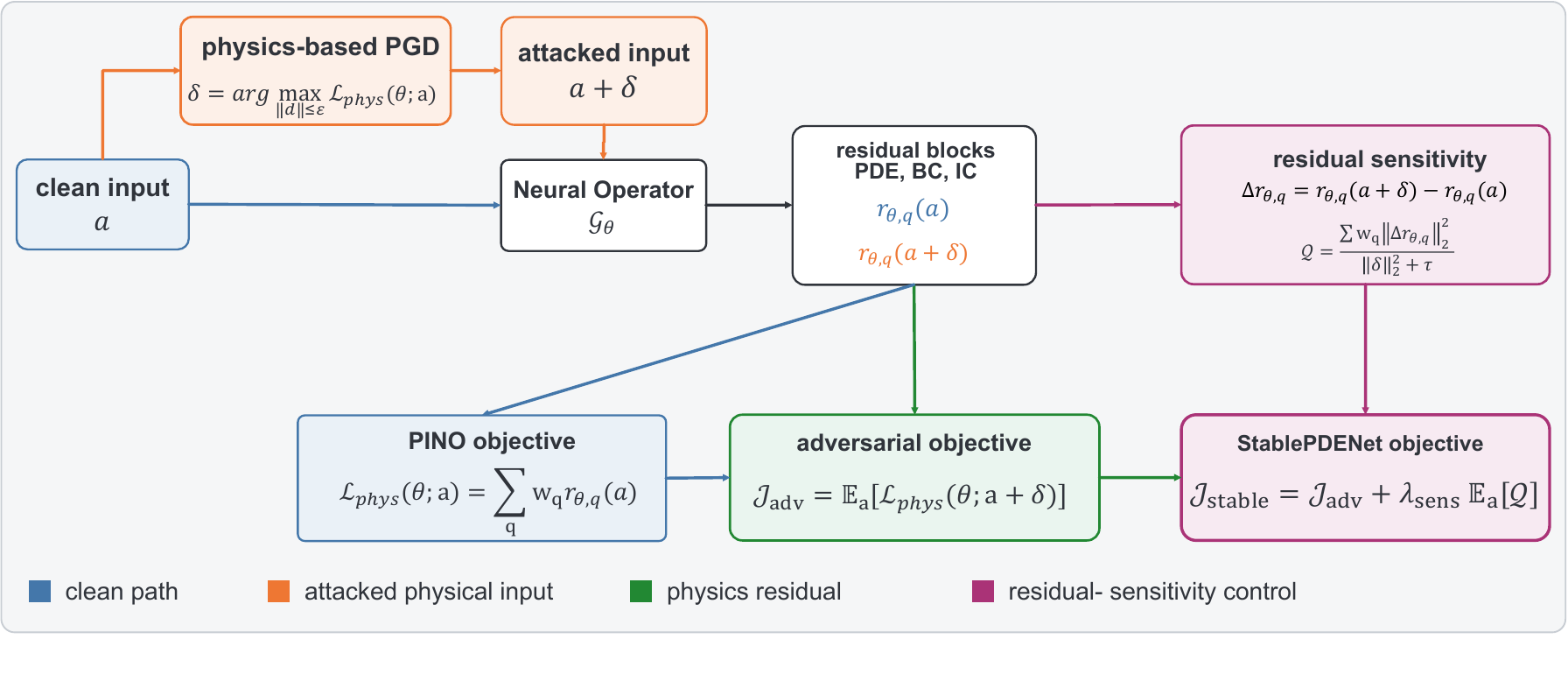}
	 \caption{Architecture of StablePDENet. A physics-based PGD adversary
		constructs $a+\delta$ by maximizing the attacked physics loss. The
		clean and attacked inputs are evaluated by the same neural operator, and their
		residual-vector difference forms the normalized residual-sensitivity penalty. The
		method is physics-informed and does not require paired solution labels during
		training.}
	\label{fig:stablepdenet-architecture}
\end{figure}

Figure~\ref{fig:stablepdenet-architecture} summarizes the physics-informed
operator and the proposed training objective. It shows the clean and
attacked forward passes used to construct the physics-informed objective, the adversarial objective and the
normalized residual-sensitivity penalty.

\subsection{\texorpdfstring{Attack learning}{Attack learning}}
\label{subsec:attack-learning}
Following the projected-gradient framework of
Madry et al.~\cite{madry2018towards}, we construct an input perturbation by
maximizing a differentiable attack objective over a prescribed admissible input space $\mathcal{A}$.
During training, the attack objective is the physics loss,
\begin{equation}
	\delta_\theta
	\approx
	\arg\max_{\delta\in\mathcal{U}(a;\varepsilon)}
	\mathcal{L}_{\rm phys}(\theta;a+\delta),
	\label{eq:training-attack}
\end{equation}
where $\mathcal{U}(a;\varepsilon)$ imposes both physical admissibility and the
selected perturbation budget. At evaluation, the same projected-gradient
updates are used with the perturbation objective. We consider the following two choices of
$\mathcal{U}(a;\varepsilon)$.

\paragraph{Pointwise PGD-$\ell_\infty$.}
For a discretized input, define
\begin{equation}
	\mathcal{C}_\infty(a;\varepsilon)
	:=
	\left\{
	\widetilde a\in\mathcal{A}:
	\|\widetilde a-a\|_\infty\leq\varepsilon
	\right\}.
	\label{eq:linf-admissible-set}
\end{equation}
Starting from $\widetilde a^{(0)}\in\mathcal{C}_\infty(a;\varepsilon)$,
the projected ascent step is
\begin{equation}
	\widetilde a^{(k+1)}
	=
	\Pi_{\mathcal{C}_\infty(a;\varepsilon)}
	\left[
	\widetilde a^{(k)}
	+
	\alpha\,
	\operatorname{sign}\!\left(
	\nabla_{\widetilde a}\mathcal{L}_{\rm phys}
	(\theta; \widetilde a^{(k)})
	\right)
	\right].
	\label{eq:pgd-attack-linf}
\end{equation}
This constraint bounds every sampled input component and is used as a
conservative pointwise stress test. It may admit dense or grid-scale
perturbations and is therefore not intended to represent smooth measurement
noise.

\paragraph{Band-limited PGD-$\ell_2$.}
Let $\{\phi_n\}_{n=1}^{K_{\rm adv}}$ be prescribed resolved modes and
write the perturbation as
$\delta_\beta=\sum_{n=1}^{K_{\rm adv}}\beta_n\phi_n$. 
Let $\varepsilon>0$ denote a dimensionless relative perturbation level.
	For an input $a$ represented by the coefficient vector $\bm c$, we define the
	coefficient-space radius as
	\[
	\rho_2(a;\varepsilon)
	:=
	\varepsilon\|\bm c\|_2.
	\]
	The admissible set for band-limited PGD-$\ell_2$ is then
\begin{equation}
		\mathcal{C}_{2}(a;\varepsilon)
		:=
		\left\{
		\bm\beta\in\mathbb{R}^{K_{\rm adv}}:
		\|\bm\beta\|_2
		\leq
		\rho_2(a;\varepsilon)
		=
		\varepsilon\|\bm c\|_2,
		\quad
		a+\delta_{\bm\beta}\in\mathcal{A}
		\right\},
	\label{eq:bl-l2-admissible-set}
\end{equation}
where
	$\delta_{\bm\beta}
	=\sum_{n=1}^{K_{\rm adv}}\beta_n\phi_n$
	is the perturbation induced by the modal coefficient vector $\bm\beta$.
	Thus, $\rho_2(a;\varepsilon)$ is the radius of the admissible ball in
	coefficient space; its dependence on $a$ makes the perturbation budget relative
	to the magnitude of the input. In particular, $\varepsilon$ specifies the
	maximum relative coefficient perturbation,
	\[
	\frac{\|\bm\beta\|_2}{\|\bm c\|_2}
	\leq \varepsilon.
	\]
	When the modes $\{\phi_n\}$ are orthonormal, Parseval's identity gives
	$\|\delta_{\bm\beta}\|_{L^2}=\|\bm\beta\|_2$ and
	$\|a\|_{L^2}=\|\bm c\|_2$. Consequently, the same constraint can be written in
	physical function space as
	\[
	\|\delta_{\bm\beta}\|_{L^2}
	\leq
	\varepsilon\|a\|_{L^2}.
	\]
With
$g_\beta^{(k)}=\nabla_\beta\mathcal{L}_{\rm phys}(\theta; a+\delta_{\beta^{(k)}})$,
the coefficient update is
\begin{equation}
	\bm \beta^{(k+1)}
	=
	\Pi_{\mathcal{C}_{2}(a;\varepsilon)}
	\left[
	\bm \beta^{(k)}
	+
	\alpha
	\frac{g_{\bm \beta}^{(k)}}
	{\|g_{\bm \beta}^{(k)}\|_2+\eta_{\rm pgd}}
	\right],
	\qquad \eta_{\rm pgd}>0.
	\label{eq:pgd-attack-bl-l2}
\end{equation}
The $\ell_2$ constraint controls the perturbation energy, whereas the
restriction to $K_{\rm adv}$ resolved modes supplies the band limitation and
excludes unresolved grid-scale components. Thus, the smoothness or spatial
structure comes from the selected modal subspace rather than from the
$\ell_2$ norm alone. The benchmark-specific choice between
PGD-$\ell_\infty$ and band-limited PGD-$\ell_2$ is reported in
Table~\ref{tab:benchmark-setup}.

\subsection{Residual-based physics losses}
For either attack defined in Section~\ref{subsec:attack-learning}, the
perturbation acts on the physical input: every residual block that contains the
input is evaluated consistently at $a+\delta$.
The clean PI-DeepONet minimizes \eqref{eq:physics-loss}. The adversarial baseline, named AdvPI, additionally minimizes the attacked physics loss,
\begin{equation}
	\mathcal{J}_{\rm Adv}(\theta)
	=\E_a\left[
	\mathcal{L}_{\rm phys}(\theta;a+\delta_\theta)
	\right].
	\label{eq:adv-objective}
\end{equation}
The implementation uses the equation-specific block weights in both the inner attack and attacked outer loss.

%\paragraph{\rev{Discrete norms used in the implementation.}}
The theoretical estimates use continuous function-space norms, whereas
the implementation evaluates inputs and residuals at finitely many sensor and
collocation points. For a sampled vector
$\mathbf v=(v_1,\ldots,v_N)$, we use the normalized discrete $\ell_2$ norm
\begin{equation}
	\|\mathbf v\|_{2,N}^{2}
	:=
	\frac{1}{N}
	\sum_{\ell=1}^{N}\|v_\ell\|_2^2.
	\label{eq:normalized-discrete-l2}
\end{equation}
For residual block $q\in\mathcal{B}$, collect the sampled change caused
by the input perturbation in the vector
\begin{equation}
	\Delta\mathbf r_{\theta,q}(a,\delta)
	:=
	\left(
	r_{\theta,q}(a+\delta;\eta_{q,\ell})
	-
	r_{\theta,q}(a;\eta_{q,\ell})
	\right)_{\ell=1}^{N_q}.
	\label{eq:sampled-residual-response}
\end{equation}
The factor $1/N_q$ in
$\|\Delta\mathbf r_{\theta,q}\|_{2,N_q}^{2}$ prevents a residual block from
gaining weight solely because it uses more collocation points, while $w_q$
specifies the relative contribution of block $q$. For an input perturbation
$\delta\in\mathbb{R}^{m}$, the same convention gives
$\|\delta\|_{2,m}^{2}=m^{-1}\|\delta\|_2^2$. These sampled quantities are
numerical approximations of the corresponding continuous norms; their accuracy
depends on the sampling rule and resolution.

Using this convention, define the residual-sensitivity quotient by
\begin{equation}
	\mathcal{Q}_{\mathcal{B}}(\theta;a,\delta)
	=\frac{
		\displaystyle\sum_{q\in\mathcal{B}}w_q
		\left\|\Delta\mathbf r_{\theta,q}(a,\delta)\right\|_{2,N_q}^{2}
	}{
		\|\delta\|_{2,m}^{2}+\tau
	},
	\qquad \tau=10^{-6}.
	\label{eq:increment-loss}
\end{equation}
This is a difference of residual vectors, not a difference of scalar physics losses. 
Writing $r_\theta(a):= \mathcal{F}(\mathcal{G}_\theta(a),a)$, the term \eqref{eq:increment-loss} is a finite-difference approximation of the
Fr\'echet derivative of the residual map,
\begin{equation}
	r_{\theta}(a+\delta)-r_{\theta}(a)
	=
	D_{a}r_{\theta}(a)\left[\delta\right]+o(\|\delta\|) ,
\end{equation}
so that $\mathcal{Q}_{\mathcal{B}}$ penalizes the
the directional linearized residual sensitivity along the sampled perturbations rather than the residual magnitude itself.
Whereas $\mathcal{L}_{\mathrm{phys}}$ only asks the network to remain
physically consistent at a perturbed input, $\mathcal{Q}_{\mathcal{B}}$
asks the sensitivity of the prediction to be physically consistent as
well.

StablePDENet minimizes
\begin{equation}
	\mathcal{J}_{\rm Stable}(\theta)
	=\mathcal{J}_{\rm Adv}(\theta)
	+\lambda_{\rm sens}\,
	\E_a\left[\mathcal{Q}_{\mathcal{B}}
	(\theta;a,\delta_\theta)\right].
	\label{eq:stable-objective}
\end{equation}
where a small regularization parameter $\tau>0$ prevents division by zero.
$\lambda_{\rm sens}$ is calibrated on attacked warm-up batches to an initial target share $\rho=0.10$,
\begin{equation}
	\lambda_{\rm sens}
	=\frac{\rho}{1-\rho}
	\frac{\overline{\mathcal{L}}_{\rm phys}}
	{\overline{\mathcal{Q}}_{\mathcal{B}}},
	\label{eq:lambda-calibration}
\end{equation}
where $\overline{\mathcal{L}}_{\rm phys}$ and $\overline{\mathcal{Q}}_{\mathcal{B}}$ represent the average magnitudes of the physics loss and the residual-sensitivity quotient, respectively. The calibration target is not a claim that the regularizer retains a $10\%$ share throughout training.

\subsection{Residual Sensitivity and Error-Operator Stability}
A frequently used surrogate for robustness is the direct penalization of the
prediction increment,
\begin{equation}
	\left\|
	\mathcal{G}_{\theta}(a+\delta)-\mathcal{G}_{\theta}(a)
	\right\| .
	\label{eq:naive-robustness}
\end{equation}
This is inappropriate whenever the underlying PDE possesses a genuine,
physically meaningful sensitivity: near-resonant, low-viscosity or
high-frequency problems all exhibit large but correct amplification of small
input perturbations. Driving \eqref{eq:naive-robustness} to zero therefore
trades model-induced instability for a systematic loss of physical fidelity.
The property we actually require is the consistency of the perturbation
response,
\begin{equation}
	\mathcal{G}_{\theta}(a+\delta)-\mathcal{G}_{\theta}(a)
	\;\approx\;
	\mathcal{S}(a+\delta)-\mathcal{S}(a) .
	\label{eq:response-consistency}
\end{equation}
Defining the error operator
\begin{equation}
	\mathcal{E}_\theta(a)=\Gop_\theta(a)-\Sop(a),
	\label{eq:error-operator}
\end{equation}
gives the identity
\begin{equation}
	\Gop_\theta(a+\delta)-\Gop_\theta(a)
	=\Sop(a+\delta)-\Sop(a)
	+\mathcal{E}_\theta(a+\delta)-\mathcal{E}_\theta(a).
	\label{eq:response-decomposition}
\end{equation}

Suppose the true solution operator is Lipschitz on a set $\mathcal{D}$,
\begin{equation}\label{eq:true_stability}
	\norm{\Sop(a_2)-\Sop(a_1)}_{\mathcal{Y}}
	\leq C_{\Sop}\norm{a_2-a_1}_{\mathcal{X}}.
\end{equation}
If the error operator \eqref{eq:error-operator} is Lipschitz with constant $C_{\mathcal{E}}$, then \eqref{eq:response-decomposition} yields
\begin{equation}
	\norm{\Gop_\theta(a_2)-\Gop_\theta(a_1)}_{\mathcal{Y}}
	\leq (C_{\Sop}+C_{\mathcal{E}})
	\norm{a_2-a_1}_{\mathcal{X}}.
	\label{eq:learned-lipschitz}
\end{equation}
The objective is therefore to reduce the sensitivity of the error operator,
as quantified by $C_{\mathcal E}$, rather than to minimize the overall input
sensitivity $L_\theta$ of the learned operator.

\subsubsection{Linear Residual-to-Error Stability Bound}

Consider a linear source-to-solution problem
\begin{equation}
	A u=Ba,
	\qquad \Sop=A^{-1}B,
	\label{eq:linear-pde}
\end{equation}
	in compatible spaces. When boundary or initial conditions are imposed
	softly, $A:\mathcal{Y}\to\mathcal{Z}$ denotes the stacked operator containing both the interior PDE equation and all softly enforced auxiliary conditions, such as boundary or initial conditions. Here, $A^{-1}$ means a bounded
	left inverse on the compatible range of this augmented operator. Bounded
	inverse stability is an explicit analytical assumption on the residual norm
	used in the theorem. The learned residual is
\begin{equation}
	r_\theta(a)=A\Gop_\theta(a)-Ba.
\end{equation}

\begin{theorem}[Linear residual-sensitivity control]\label{the:linear}
	Assume that $A^{-1}$ is bounded and that
	\begin{equation}
		\norm{r_\theta(a_2)-r_\theta(a_1)}_{\mathcal{Z}}
		\leq \gamma_r\norm{a_2-a_1}_{\mathcal{X}}
	\end{equation}
	on $\mathcal{D}$. Then
	\begin{equation}
		\norm{\mathcal{E}_\theta(a_2)-\mathcal{E}_\theta(a_1)}_{\mathcal{Y}}
		\leq \norm{A^{-1}}_{\mathcal{Z}\to\mathcal{Y}}
		\gamma_r\norm{a_2-a_1}_{\mathcal{X}},
		\label{eq:linear-error-bound}
	\end{equation}
	and the learned operator satisfies \eqref{eq:learned-lipschitz} with
	$C_{\mathcal{E}}\leq\norm{A^{-1}}\gamma_r$.
\end{theorem}

\begin{proof}
	Since $A\Sop(a)=Ba$, one has $r_\theta(a)=A\mathcal{E}_\theta(a)$. Hence
	\begin{equation}
		\mathcal{E}_\theta(a_2)-\mathcal{E}_\theta(a_1)
		=A^{-1}\bigl(r_\theta(a_2)-r_\theta(a_1)\bigr),
	\end{equation}
	and taking norms proves the result.
\end{proof}
The denominator in \eqref{eq:increment-loss} makes its finite-difference
value an empirical squared residual slope along the sampled perturbation.

\subsubsection{Local Nonlinear Residual-to-Error Stability Bound}

For a nonlinear PDE, the identity
$r_\theta(a)=A\mathcal E_\theta(a)$ used in the linear case is no longer
available with a fixed operator $A$. Nevertheless, a local analogue
can be obtained by replacing $A$ with an averaged linearization of the residual operator along the segment joining the exact and learned
solutions. Here, the residual operator $\mathcal{F}$ includes the PDE,
boundary-condition, and initial-condition blocks whenever they are
present.

\begin{theorem}[Local nonlinear residual-to-error and residual-sensitivity control]
	Let $\mathcal{D}\subset \mathcal X$, and let
	$\mathcal{F}:\mathcal Y\times \mathcal X\rightarrow \mathcal Z$ be continuously Fr\'echet
	differentiable with respect to its first argument. Suppose that
	$\mathcal{F}(\Sop(a),a)=0$ for every $a\in\mathcal{D}$ and that the
	true solution operator satisfies \eqref{eq:true_stability}.
	
	Assume that the linearized residual operator at the exact solution
	is uniformly bounded from below:
	\begin{equation}
		\left\|
		D_u\mathcal{F}(\Sop(a),a)[v]
		\right\|_{\mathcal Z}
		\geq
		\beta\|v\|_{\mathcal Y},
		\qquad
		a\in\mathcal{D},\quad v\in \mathcal Y,
		\label{eq:nonlinear_inverse_stability}
	\end{equation}
	for some $\beta>0$.
	
	Let
	\begin{equation}
		\mathcal{T}_{\rho_E}
		=
		\left\{
		(v,a)\in \mathcal Y\times\mathcal{D}:
		\|v-\Sop(a)\|_{\mathcal Y}\leq\rho_E
		\right\}.
	\end{equation}
	Suppose that, on $\mathcal{T}_{\rho_E}$,
	\begin{equation}
		\begin{aligned}
			&
			\left\|
			D_u\mathcal{F}(v_2,a_2)
			-
			D_u\mathcal{F}(v_1,a_1)
			\right\|_{\mathcal{L}(\mathcal Y,\mathcal Z)}
			\\
			&\qquad\leq
			L_u\|v_2-v_1\|_{\mathcal Y}
			+
			L_a\|a_2-a_1\|_{\mathcal X},
		\end{aligned}
		\label{eq:nonlinear_jacobian_lipschitz}
	\end{equation}
	and that the learned operator remains in this tube:
	\begin{equation}
		\sup_{a\in\mathcal{D}}
		\|\mathcal E_\theta(a)\|_{\mathcal Y}
		=
		\sup_{a\in\mathcal{D}}
		\|\Gop_\theta(a)-\Sop(a)\|_{\mathcal Y}
		\leq
		\rho_E.
		\label{eq:error_tube}
	\end{equation}
	If
	\begin{equation}
		\beta-L_u\rho_E>0,
		\label{eq:nonlinear_tube_condition}
	\end{equation}
	then the following estimates hold.
	
	First, for every $a\in\mathcal{D}$,
	\begin{equation}
		\|\mathcal E_\theta(a)\|_{\mathcal Y}
		\leq
		\frac{
			\|r_\theta(a)\|_{\mathcal Z}
		}{
			\beta-\frac{L_u\rho_E}{2}
		}.
		\label{eq:nonlinear_residual_to_error}
	\end{equation}
	
	Second, for every $a_1,a_2\in\mathcal{D}$,
	\begin{equation}
		\begin{aligned}
			&
			\|\mathcal E_\theta(a_2)-\mathcal E_\theta(a_1)\|_{\mathcal Y}
			\\
			&\quad\leq
			\frac{
				\|r_\theta(a_2)-r_\theta(a_1)\|_{\mathcal Z}
				+
				\rho_E
				(L_uC_{\Sop}+L_a)
				\|a_2-a_1\|_{\mathcal X}
			}{
				\beta-L_u\rho_E
			}.
		\end{aligned}
		\label{eq:nonlinear_residual_sensitivity_bound}
	\end{equation}
	
	In particular, if
	\begin{equation}
		\|r_\theta(a_2)-r_\theta(a_1)\|_{\mathcal Z}
		\leq
		\gamma_r\|a_2-a_1\|_{\mathcal X},
		\label{eq:nonlinear_residual_lipschitz}
	\end{equation}
	then the error operator is locally Lipschitz:
	\begin{equation}
		\|\mathcal E_\theta(a_2)-\mathcal E_\theta(a_1)\|_{\mathcal Y}
		\leq
		\gamma_E\|a_2-a_1\|_{\mathcal X},
		\label{eq:nonlinear_error_lipschitz}
	\end{equation}
	where
	\begin{equation}
		\gamma_E
		=
		\frac{
			\gamma_r
			+
			\rho_E(L_uC_{\Sop}+L_a)
		}{
			\beta-L_u\rho_E
		}.
		\label{eq:nonlinear_error_constant}
	\end{equation}
	Consequently,
	\begin{equation}
		\|\Gop_\theta(a_2)-\Gop_\theta(a_1)\|_{\mathcal Y}
		\leq
		(C_{\Sop}+\gamma_E)
		\|a_2-a_1\|_{\mathcal X}.
		\label{eq:nonlinear_learned_operator_stability}
	\end{equation}
\end{theorem}

\begin{proof}
	For $i=1,2$, define
	\begin{equation}
		u_i=\Sop(a_i),
		\qquad
		e_i=\mathcal E_\theta(a_i),
		\qquad
		r_i=r_\theta(a_i).
	\end{equation}
	Since
	$\mathcal{F}(u_i,a_i)=0$ and
	$\Gop_\theta(a_i)=u_i+e_i$, the fundamental theorem of calculus
	in Banach spaces gives
	\begin{equation}
		\begin{aligned}
			r_i
			&=
			\mathcal{F}(u_i+e_i,a_i)
			-
			\mathcal{F}(u_i,a_i)
			\\
			&=
			\left[
			\int_0^1
			D_u\mathcal{F}(u_i+te_i,a_i)\,dt
			\right]e_i.
		\end{aligned}
		\label{eq:averaged_linearization_identity}
	\end{equation}
	Define the averaged linearized operator
	\begin{equation}
		\overline{A}_i
		=
		\int_0^1
		D_u\mathcal{F}(u_i+te_i,a_i)\,dt.
		\label{eq:averaged_linearized_operator}
	\end{equation}
	Then
	\begin{equation}
		r_i=\overline{A}_ie_i.
		\label{eq:nonlinear_residual_error_identity}
	\end{equation}
	
	For every $v\in \mathcal Y$, equations
	\eqref{eq:nonlinear_inverse_stability},
	\eqref{eq:nonlinear_jacobian_lipschitz}, and
	\eqref{eq:error_tube} imply
	\begin{equation}
		\begin{aligned}
			\|\overline{A}_iv\|_{\mathcal Z}
			&\geq
			\left\|
			D_u\mathcal{F}(u_i,a_i)[v]
			\right\|_{\mathcal Z}
			\\
			&\quad-
			\int_0^1
			\left\|
			\left[
			D_u\mathcal{F}(u_i+te_i,a_i)
			-
			D_u\mathcal{F}(u_i,a_i)
			\right][v]
			\right\|_{\mathcal Z}\,dt
			\\
			&\geq
			\beta\|v\|_{\mathcal Y}
			-
			L_u
			\int_0^1
			t\|e_i\|_{\mathcal Y}\|v\|_{\mathcal Y}\,dt
			\\
			&\geq
			\left(
			\beta-\frac{L_u\rho_E}{2}
			\right)
			\|v\|_{\mathcal Y}.
		\end{aligned}
		\label{eq:averaged_operator_lower_bound}
	\end{equation}
	Taking $v=e_i$ and using
	$r_i=\overline{A}_ie_i$ proves
	\eqref{eq:nonlinear_residual_to_error}.
	
	We next consider the residual sensitivity. From
	\eqref{eq:nonlinear_residual_error_identity},
	\begin{equation}
		r_2-r_1
		=
		\overline{A}_2(e_2-e_1)
		+
		(\overline{A}_2-\overline{A}_1)e_1.
		\label{eq:nonlinear_response_decomposition}
	\end{equation}
	Let
	\begin{equation}
		d=\|a_2-a_1\|_{\mathcal X}.
	\end{equation}
	Using
	\eqref{eq:nonlinear_jacobian_lipschitz} and
	\eqref{eq:true_stability}, we obtain
	\begin{equation}
		\begin{aligned}
			&
			\|\overline{A}_2-\overline{A}_1\|_{\mathcal{L}(\mathcal Y,\mathcal Z)}
			\\
			&\leq
			\int_0^1
			\Bigl[
			L_u
			\|u_2-u_1+t(e_2-e_1)\|_{\mathcal Y}
			+
			L_a d
			\Bigr]dt
			\\
			&\leq
			(L_uC_{\Sop}+L_a)d
			+
			\frac{L_u}{2}
			\|e_2-e_1\|_{\mathcal Y}.
		\end{aligned}
		\label{eq:averaged_operator_difference}
	\end{equation}
	Combining
	\eqref{eq:averaged_operator_lower_bound},
	\eqref{eq:nonlinear_response_decomposition}, and
	\eqref{eq:averaged_operator_difference} yields
	\begin{equation}
		\begin{aligned}
			\left(
			\beta-\frac{L_u\rho_E}{2}
			\right)
			\|e_2-e_1\|_{\mathcal Y}
			&\leq
			\|r_2-r_1\|_{\mathcal Z}
			\\
			&\quad+
			\rho_E(L_uC_{\Sop}+L_a)d
			\\
			&\quad+
			\frac{L_u\rho_E}{2}
			\|e_2-e_1\|_{\mathcal Y}.
		\end{aligned}
		\label{eq:nonlinear_response_intermediate}
	\end{equation}
	Moving the final term to the left-hand side gives
	\begin{equation}
		(\beta-L_u\rho_E)
		\|e_2-e_1\|_{\mathcal Y}
		\leq
		\|r_2-r_1\|_{\mathcal Z}
		+
		\rho_E(L_uC_{\Sop}+L_a)d.
	\end{equation}
	Because $\beta-L_u\rho_E>0$, division by this quantity proves
	\eqref{eq:nonlinear_residual_sensitivity_bound}.
	
	Substituting
	\eqref{eq:nonlinear_residual_lipschitz} into
	\eqref{eq:nonlinear_residual_sensitivity_bound} gives
	\eqref{eq:nonlinear_error_lipschitz}--%
	\eqref{eq:nonlinear_error_constant}. Finally,
	\begin{equation}
		\begin{aligned}
			\Gop_\theta(a_2)-\Gop_\theta(a_1)
			={}&
			\Sop(a_2)-\Sop(a_1)
			\\
			&+
			\mathcal E_\theta(a_2)-\mathcal E_\theta(a_1),
		\end{aligned}
	\end{equation}
	and the triangle inequality together with
	\eqref{eq:true_stability} proves
	\eqref{eq:nonlinear_learned_operator_stability}.
\end{proof}

\begin{remark}[Scope of the analysis]
	The theorems provide conditional a priori residual-to-response estimates. In practice, $\mathcal{Q}_{\mathcal{B}}$ evaluates residual sensitivity only over finitely many samples, perturbation radii, collocation points, and adversarial directions using normalized discrete $\ell_2$ norms. Because the required norm equivalence and nonlinear stability assumptions are not verified a posteriori, the analysis motivates the method but does not provide a global stability certificate for the trained model.
\end{remark}

\subsection{Training algorithm for Attack and Defense}

Overall, Algorithms~\ref{alg:stable_attack} and
\ref{alg:stable_train} summarize the inner adversarial search and the
outer optimization of StablePDENet, respectively.
With the model parameters fixed, Algorithm~\ref{alg:stable_attack}
constructs an adversarial input by maximizing the physics-informed
loss over the prescribed perturbation set. It implements the two
projected-gradient constructions defined in
Section~\ref{subsec:attack-learning}. Algorithm~\ref{alg:stable_train} first performs a clean-training warm-up and then switches to cached
adversarial defense updates.  The integer $M$ denotes the refresh interval
of the cached adversarial batch. In most experiments, the warm-up lasts $j_0=5000$ steps.  A
new adversarial batch is then generated every $M=1000$ steps using
$n_{\mathrm{iter}}=40$ PGD iterations and is reused during the following
$M$ outer optimization steps.  Thus, every post-warm-up outer step minimizes
the attacked physics loss together with the residual-sensitivity penalty.

The physics-informed objective is embedded in both the inner and outer
optimization problems. In the inner problem, PGD searches for an
admissible input perturbation that maximizes the PDE, boundary, and
initial-condition residuals. In the outer problem, the same
adversarial input is used to evaluate both the attacked physics loss
and the residual-sensitivity regularizer. Therefore, only one
adversarial batch needs to be generated at each cache refresh, and that
	batch is reused for the subsequent $M$ defense updates.
Moreover, because the attack is driven directly by differentiable
physics residuals, generating a perturbed input does not require
computing a new reference PDE solution at every PGD iteration.
StablePDENet thereby provides an efficient inner--outer framework for
jointly maintaining nominal physical consistency and controlling the
sensitivity of the learned operator to admissible input perturbations.
\begin{algorithm}[H]
	\caption{Physics-based PGD attack for StablePDENet}
	\label{alg:stable_attack}
	\small
	\begin{algorithmic}
		\setlength{\itemsep}{0pt}
		\setlength{\parsep}{0pt}
		
		\REQUIRE Clean input $a$, coordinates $\xi$, fixed operator
		$\mathcal{G}_{\theta}$, attack type $T_{\mathrm{adv}}$,
		step size $\alpha$, perturbation bound $\varepsilon$,
		iterations $n_{\mathrm{iter}}$, and admissible set $\mathcal{A}$;
		for BL-PGD-$\ell_2$, eigenmodes
		$\{\phi_n\}_{n=1}^{K_{\mathrm{adv}}}$ and
		$\eta_{\mathrm{pgd}}>0$
		
		\IF{$T_{\mathrm{adv}}
			=\mathrm{PGD}\mbox{-}\ell_\infty$}
		
		\STATE
		$\delta^{(0)}
		\sim\mathcal{U}(-\varepsilon,\varepsilon)$,
		$\tilde a^{(0)}
		\gets
		\Pi_{\mathcal{C}_{\infty}(a;\varepsilon)}
		\left(a+\delta^{(0)}\right)$
		
		\ELSE
		
		\STATE
		$\bm z\sim\mathcal{N}(\bm 0,I)$,
		$\bm\beta^{(0)}
		\gets
		\Pi_{\mathcal{C}_{2}(a; \varepsilon)}(\bm z)$,
		$\tilde a^{(0)}
		\gets
		a+\sum_{n=1}^{K_{adv}}\beta_n^{(0)}\phi_n$
		
		\ENDIF
		
		\FOR{$r=0$ to $n_{\mathrm{iter}}-1$}
		
		\IF{$T_{\mathrm{adv}}
			=\mathrm{PGD}\mbox{-}\ell_\infty$}
		
		\STATE
		$g^{(r)}
		\gets
		\nabla_{\tilde a}
		\mathcal{L}_{\mathrm{phys}}
		\left(
		\theta;\tilde a^{(r)},\xi
		\right)$
		
		\STATE
		$\tilde a^{(r+1)}
		\gets
		\Pi_{\mathcal{C}_{\infty}(a;\varepsilon)}
		\left[
		\tilde a^{(r)}
		+
		\alpha\,
		\operatorname{sign}
		\left(g^{(r)}\right)
		\right]$
		
		\ELSE
		
		\STATE
		$g_{\bm \beta}^{(r)}
		\gets
		\nabla_{\bm\beta}
		\mathcal{L}_{\mathrm{phys}}
		\left(
		\theta;
		a+\sum_{n=1}^{K_{adv}}\beta_n^{(r)}\phi_n,
		\xi
		\right)$
		
		\STATE
		$\bm\beta^{(r+1)}
		\gets
		\Pi_{\mathcal{C}_{2}(a;\varepsilon)}
		\left[
		\bm\beta^{(r)}
		+
		\alpha
		\frac{g_{\bm \beta}^{(r)}}
		{
			\|g_{\bm \beta}^{(r)}\|_2
			+
			\eta_{\mathrm{pgd}}
		}
		\right]$
		
		\STATE
		$\tilde a^{(r+1)}
		\gets
		a+\sum_{n=1}^{K_{adv}}\beta_n^{(r+1)}\phi_n$
		
		\ENDIF
		
		\ENDFOR
		
		\STATE
		$\delta^{\mathrm{adv}}
		\gets
		\tilde a^{(n_{\mathrm{iter}})}-a$
		
		\RETURN
		$(\tilde a^{(n_{\mathrm{iter}})},
		\delta^{\mathrm{adv}})$
		
	\end{algorithmic}
	\vspace{-0.4em}
\end{algorithm}

\begin{algorithm}[H]
	\caption{StablePDENet training}
	\label{alg:stable_train}
	\small
		\begin{algorithmic}
			\setlength{\itemsep}{0pt}
			\setlength{\parsep}{0pt}
			
			\REQUIRE Initial parameters $\theta_0$, training steps $N$,
			batch size $B$, learning rate $\eta_{\theta}$,
			warm-up step $j_0$, attack-refresh interval $M$,
			attack parameters, residual-sensitivity coefficient
			$\lambda_{\mathrm{sens}}$, weights
			$\{w_q\}_{q\in\mathcal{B}}$,
			and $\tau>0$
			
			\STATE Initialize adversarial cache
			$\mathcal{C}_{\mathrm{adv}}\gets\varnothing$
			
			\FOR{$j=0$ to $N-1$}
			
			\IF{$j<j_0$}
			
			\STATE Sample a clean batch
			$\{(a_b,\xi_b)\}_{b=1}^{B}$
			
			\STATE
			$\mathcal{J}_j
			\gets
			B^{-1}
			\sum_b
			\mathcal{L}_{\mathrm{phys}}
			\left(
			\theta_j;a_b,\xi_b
			\right)$
			
			\ELSE
			
			\IF{$\mathcal{C}_{\mathrm{adv}}=\varnothing$
				or $(j-j_0)\bmod M=0$}
			
			\STATE Sample a clean base batch
			$\{(a_b^{\mathrm c},\xi_b^{\mathrm c})\}_{b=1}^{B}$
			
			\STATE Generate
			$\{(\tilde a_b^{\mathrm c},\delta_b^{\mathrm c})\}_{b=1}^{B}$
			from $\{(a_b^{\mathrm c},\xi_b^{\mathrm c})\}_{b=1}^{B}$
			using Algorithm~\ref{alg:stable_attack}
			
			\STATE Store and detach
			$\mathcal{C}_{\mathrm{adv}}
			\gets
			\{(a_b^{\mathrm c},\xi_b^{\mathrm c},
			\tilde a_b^{\mathrm c},\delta_b^{\mathrm c})\}_{b=1}^{B}$
			
			\ENDIF
			
			\STATE Read
			$\{(a_b^{\mathrm c},\xi_b^{\mathrm c},
			\tilde a_b^{\mathrm c},\delta_b^{\mathrm c})\}_{b=1}^{B}$
			from $\mathcal{C}_{\mathrm{adv}}$
			
			\STATE
			$\mathcal{J}_{\mathrm{adv}}
			\gets
			B^{-1}
			\sum_b
			\mathcal{L}_{\mathrm{phys}}
			\left(
			\theta_j;\tilde a_b^{\mathrm c},\xi_b^{\mathrm c}
			\right)$
			
			\STATE
			$\mathcal{J}_{\mathrm{sens}}
			\gets
			B^{-1}
			\sum_b
			\mathcal{Q}_{\mathcal{B}}
			\left(\theta_j;a_b^{\mathrm c},\delta_b^{\mathrm c}\right)$
			
			\STATE
			$\mathcal{J}_j
			\gets
			\mathcal{J}_{\mathrm{adv}}
			+
			\lambda_{\mathrm{sens}}
			\mathcal{J}_{\mathrm{sens}}$
			
			\ENDIF
			
			\STATE
			$\theta_{j+1}
			\gets
			\theta_j
			-
			\eta_{\theta}
			\nabla_{\theta}\mathcal{J}_j$
			
			\ENDFOR
			
			\RETURN $\mathcal{G}_{\theta_N}$
			
		\end{algorithmic}
	\vspace{-0.4em}
\end{algorithm}

\section{Numerical Experiments}\label{section:03}
In this section, we conduct a set of numerical experiments to show the practical performance of the StablePDENet in solving parametric PDEs with perturbed inputs. 

\subsection{Experimental protocol}

\subsubsection{Evaluation attack and metrics}

The clean test set is used to evaluate nominal prediction accuracy. For each
test input, a common perturbation $\delta_{\mathrm{com}}$ is generated and
applied unchanged to PI-DeepONet, AdvPI, and StablePDENet. The three models are
therefore evaluated using the same perturbed input
$a+\delta_{\mathrm{com}}$ and the same recomputed reference solution
$\Sop(a+\delta_{\mathrm{com}})$.

For benchmark $p$, let $\mathcal{U}_p(a;\varepsilon)$ denote the corresponding
admissible perturbation set defined using one of the two constructions in
Section~\ref{subsec:attack-learning}. The common perturbation is generated
against PI-DeepONet by maximizing the clean-reference discrepancy:
\begin{equation}
	\delta_{\mathrm{com}}(a;\varepsilon)
	\approx
	\arg\max_{\delta\in\mathcal{U}_p(a;\varepsilon)}
	\MSE\left(
	\Gop_{\PI}(a+\delta)-\Sop(a)
	\right).
	\label{eq:test-attack}
\end{equation}
PI-DeepONet is selected as the attack source because it is the unregularized
baseline underlying the three methods. The attack searches for a perturbation that induces a large prediction error
in the baseline model. The same perturbation is then applied to all three models,
allowing their predictions to be compared under identical input conditions. Accordingly, this experiment is interpreted as a
common stability test, rather than as a model-specific white-box
worst-case evaluation of AdvPI or StablePDENet.
All evaluation attacks use 40 projected-gradient steps. The admissible set and
projection are benchmark dependent, while the attack objective and
common-input evaluation protocol remain unchanged. We report
$\varepsilon\in\{0.05,0.10\}$ for the attacked test sets.

\subsubsection{Benchmarks}

Table \ref{tab:benchmark-setup} shows problem settings for assessing the performance of our method across four types of parametric differential equations with their corresponding perturbed terms. Within each equation, PI-DeepONet, AdvPI, and StablePDENet use the same architecture and sampled function family. All reported results are averaged over five independent runs with different random seeds.
Within each test sample and perturbation radius, the three models are
evaluated on the identical clean input, identical perturbed input, and
identical recomputed physical reference.
The first three benchmark problems use PGD-$\ell_\infty$ attacks,
whereas the Helmholtz problem uses a band-limited PGD-$\ell_2$ attack.

\begin{table}[H]
	\centering
	\caption{Problem settings for assessing the performance of StablePDENet.}
	\label{tab:benchmark-setup}
	\small
	\renewcommand{\arraystretch}{1.5}
	\begin{tabular}{
			>{\raggedright\arraybackslash}p{1.70cm}
			>{\raggedright\arraybackslash}p{4.30cm}
			>{\raggedright\arraybackslash}p{3.35cm}
			>{\raggedright\arraybackslash}p{3.0cm}}
		\toprule
		Problem & Governing equation & Input term  \\
		\midrule
		Heat & $u_t-\alpha u_{xx}=0$ & initial condition \\
		Poisson & $-u_{xx}=f$ & source function $f$  \\
		Burgers & $u_t+uu_x=\nu u_{xx}$ & initial condition  \\
		Helmholtz & $-u_{xx}-\kappa^2u=f_c$ & sine coefficients $c$\\
		\bottomrule
	\end{tabular}
\end{table}

Generally, the networks for the branch net and trunk net are both $4$-layer fully connected neural networks with $128$ neurons per layer, and both employ tanh activation functions and are initialized using the Glorot normal distribution. 
All models are trained by Adam with $50000$ steps. In each step, the batch size is $100$.
Unless otherwise stated, the one-dimensional and time-dependent experiments use Adam with learning rate $5\times 10^{-4}$ and no learning-rate scheduler. The reported results below are averaged over $5$ runs with different random seeds.

The perturbation radius $\varepsilon$ is selected in the discrete input norm after the branch input is scaled to the range used by the solver. We choose $\varepsilon$ so that the perturbed input remains in the admissible PDE class: source and initial-condition perturbations must lead to well-posed numerical solves; coefficient perturbations are projected or rescaled to the prescribed positive range. The value of $\varepsilon$ should therefore be interpreted problem by problem rather than as a universal dimensional quantity. The interpretation of each budget follows its attack geometry in
Section~\ref{subsec:attack-learning}: $\ell_\infty$ bounds pointwise sampled
values, whereas PGD-$\ell_2$ bounds the energy of coefficients in the
prescribed modal subspace.

\subsection{\texorpdfstring{Heat Equation with Initial-Condition Perturbations}{Heat Equation with Initial-Condition Perturbations}}
\label{numsection:04}

We consider the heat equation with zero Dirichlet boundary conditions, taking the initial condition as the input function $a(x)$, 
\begin{align}
	\begin{cases}
		\frac{\partial u(x,t)}{\partial t}
		-\alpha\frac{\partial^2u(x,t)}{\partial x^2}
		=0
		\quad \text{in}\quad
		\Omega_T=\Omega\times(0,T],\\
		u(x,0)=a(x),
		\quad x\in\Omega,\\
		u(x,t)=0
		\quad \text{on}\quad
		\partial\Omega\times(0,T].
	\end{cases}
	\label{eq:heat_initial}
\end{align}
where $\Omega=[0,1]$, the diffusion coefficient is $\alpha=0.01$, and $T=1$.
The goal is to learn the operator
\[
\Sop:a\longrightarrow u(\cdot,\cdot),
\qquad
(x,t)\in[0,1]\times[0,1].
\]

\begin{figure}[htbp]
	\centering
	\includegraphics[width=\linewidth]{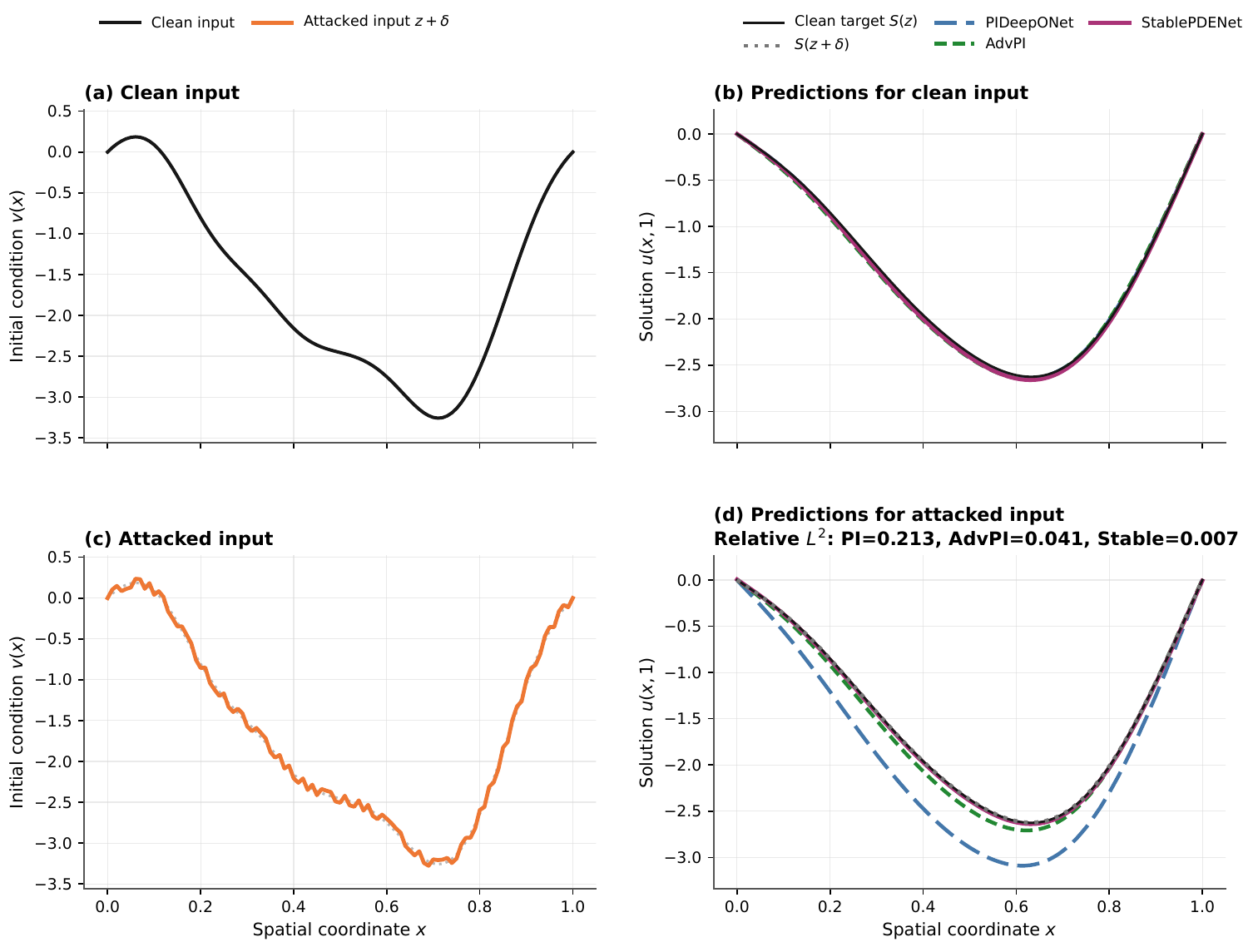}
	\caption{Heat equation with perturbed initial condition. Left: original and adversarially perturbed initial conditions; right: final-time solutions after recomputing the reference solution for the original and perturbed initial condition. The same perturbed initial condition is supplied to all three models.}
	\label{fig:heat_initial}
\end{figure}

\begin{figure}[htbp]
	\centering
	\includegraphics[width=\linewidth]{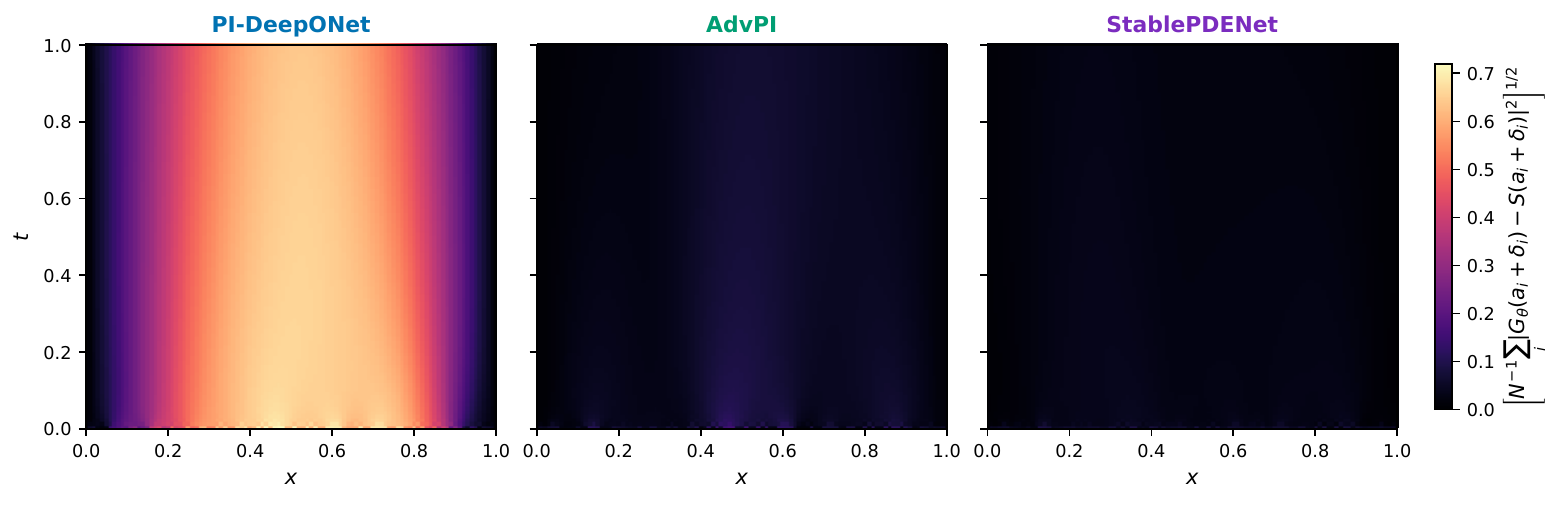}
	\caption{Space--time solution error for the heat equation with perturbed initial condition.}
	\label{fig:heat_initial_spacetime}
\end{figure}

We generate the initial-condition functions $a$ from a zero-mean Gaussian random field with unit marginal variance, a radial basis kernel and a length scale of $l=1$. $1000$ functions are evaluated on $101$ uniformly distributed sensors on $[0,1]$, along with $101$ points on the initial time $t=0$ and $40$ points on the boundaries. We use $2000$ examples to test the performance of the three models, where the test solution functions are evaluated on $[0,1]\times[0,1]$ with $101\times100$ points uniformly distributed. We compute the numerical solutions corresponding to $a$ and $\tilde{a}=a+\delta$ using the finite difference method.

\begin{figure}[htbp]
	\centering
	\includegraphics[width=\linewidth]{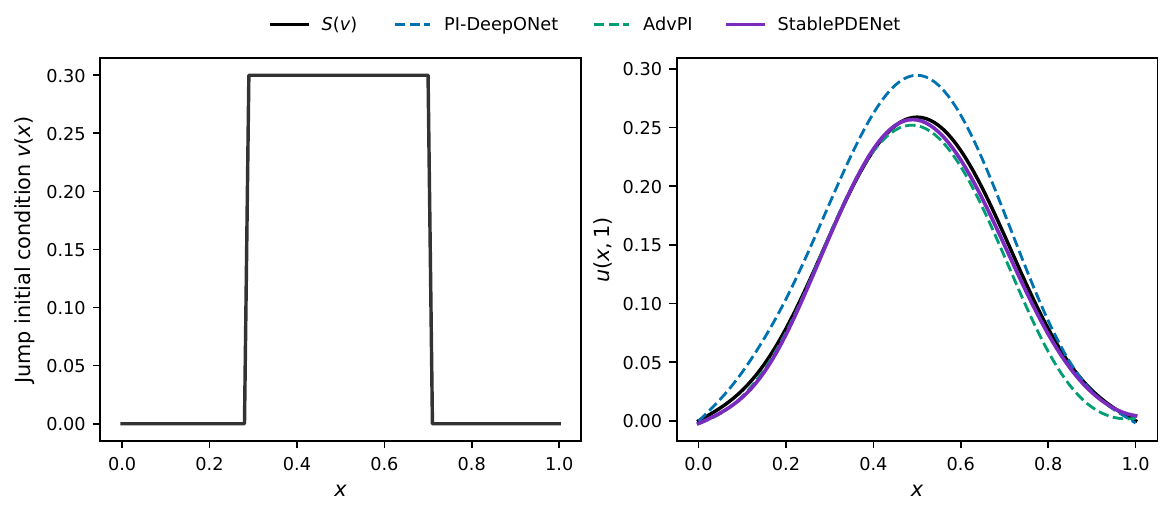}
	\caption{Heat-equation predictions for an unseen discontinuous initial condition. The model predictions are compared with the corresponding reference solution.}
	\label{fig:heat_initial_jump}
\end{figure}

\begin{figure}[htbp]
	\centering
	\includegraphics[width=\linewidth]{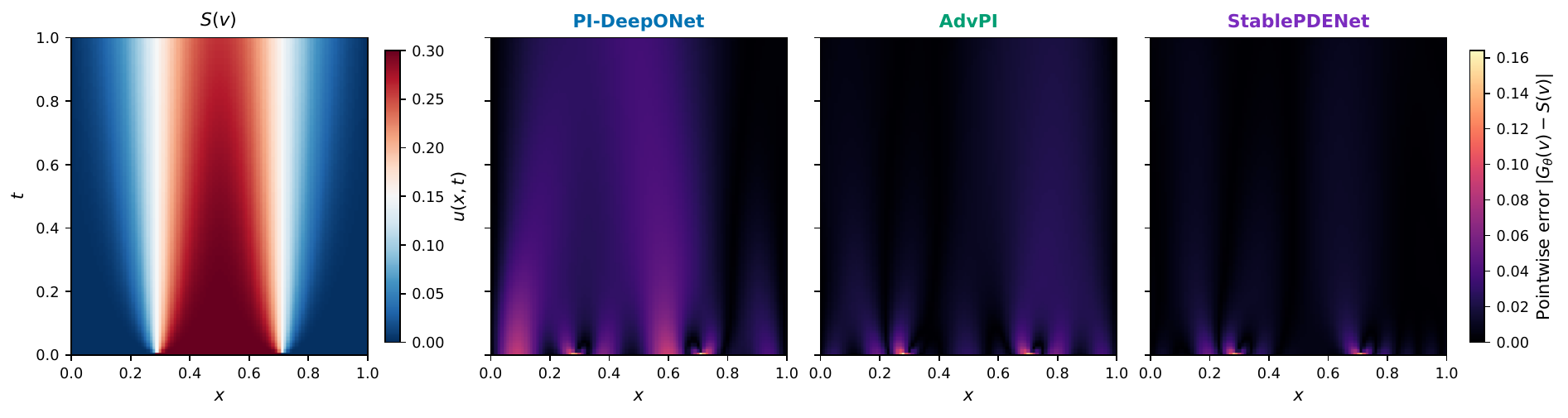}
	\caption{Space--time solution and pointwise errors for the heat equation with an unseen discontinuous initial condition.}
	\label{fig:heat_initial_spacetime_jump}
\end{figure}

Figures~\ref{fig:heat_initial} and~\ref{fig:heat_initial_spacetime} show the original initial conditions, their adversarially perturbed counterparts, and the corresponding solutions at the final-time $t=1$. The StablePDENet has lower error on the perturbed inputs, whereas PI-DeepONet is more accurate on clean inputs in this example.
The evaluation statistics are shown in Table~\ref{tab:heat_initial}. The relative error of PI-DeepONet after perturbation reaches $54\%$, while the StablePDENet remains close to its clean-input error. In addition, we evaluate the models using an initial condition with a jump
discontinuity that is not represented in the training data. The corresponding
predictions are shown in Figures~\ref{fig:heat_initial_jump}
and~\ref{fig:heat_initial_spacetime_jump}. The StablePDENet produces more accurate
predictions than the baseline models for this nonsmooth input, indicating
improved generalization beyond the training distribution.

\begin{table}[H]
	\centering
	\caption{Full space--time relative $L^2$ error for the heat equation.
		The clean row uses $\Sop(a)$ as the reference, whereas the perturbed rows use $\Sop(a+\delta_{\mathrm{com}})$.
		Bold marks the smallest entry in each row.}
	\label{tab:heat_initial}
	\small
	\begin{tabular}{lrrr}
		\toprule
		Setting & PI-DeepONet & AdvPI & StablePDENet \\
		\midrule
		Clean
		& \textbf{0.011583} & 0.019515 & 0.022877 \\
		$\varepsilon=0.05$
		& 0.540224 & 0.053075 & \textbf{0.027240} \\
		$\varepsilon=0.10$
		& 1.063989 & 0.096239 & \textbf{0.042967} \\
		\bottomrule
	\end{tabular}
\end{table}

\subsection{Poisson Equation}
\label{numsection:01}

The next equation that we try to solve is the one-dimensional Poisson equation.
\begin{align}
	\begin{cases}
		-\dfrac{\partial^2u(x)}{\partial x^2}=f(x),
		\quad x\in[0,1],\\
		u(0)=0,\\
		u(1)=0.
	\end{cases}
\end{align}
The aim is to learn the operator that maps the source function $f(x)$ to $u(x)$,
\begin{equation}
	\Sop:f(x)\longrightarrow u(x),
	\qquad x\in[0,1].
\end{equation}
The loss we use is \eqref{equ:possion_1d}.

We generate the source functions $f$ from a function space consisting of polynomials of degree $3$. $100$ functions are evaluated at $100$ sensor points uniformly distributed on $[0,1]$. The output functions are discretized on $100$ points on $[0,1]$ with a Hammersley distribution, along with two points on the boundary. We use $2000$ examples to test the performance of the three models, where the test functions are evaluated on $[0,1]$ at $100$ uniformly spaced points, and the numerical solutions corresponding to $f$ and $\tilde{f}$ are computed using the finite-difference method.

%The branch net and the trunk net are both $2$-layer fully connected networks,
%with both the hidden dimension and the output dimension being $32$.

\begin{figure}[htb]
	\centering
	\includegraphics[width=\linewidth]{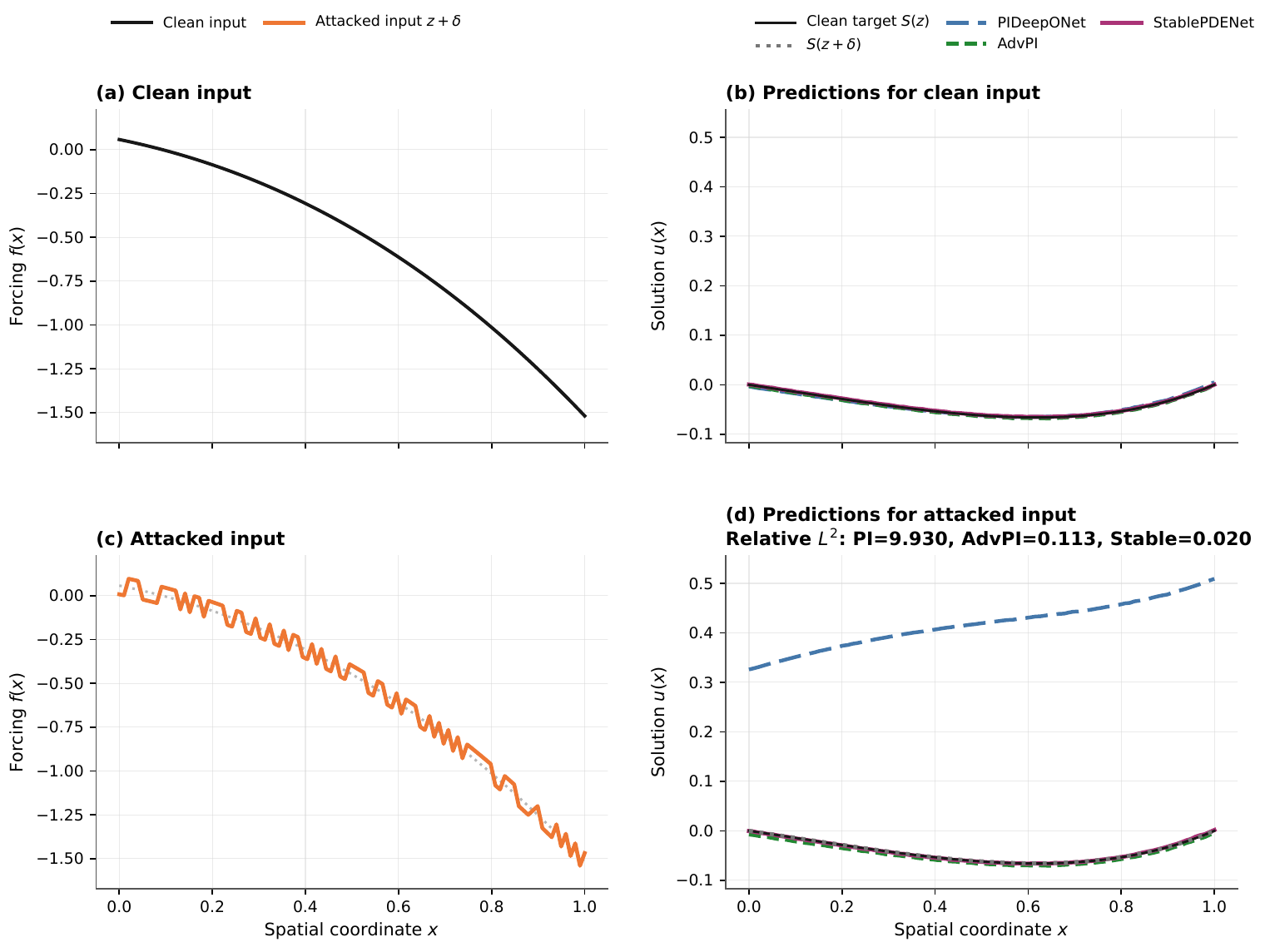}
	\caption{Poisson equation with source perturbations. Left: original and adversarially perturbed source terms; right: predictions and reference solutions for the original and perturbed sources. All three predictions correspond to the same perturbed source term.}
	\label{fig:1dpoisson}
\end{figure}

Figure~\ref{fig:1dpoisson} shows the results for a representative test sample. Similarly, the left column shows the original source term $f(x)$ and its adversarially perturbed counterpart $\tilde{f}(x)$ generated by the evaluation attack. The right column demonstrates that the three models agree well with the reference solutions for unperturbed inputs. This confirms that conventional physics-informed operator learning methods can provide accurate solutions for standard input conditions. However, our method maintains high accuracy under adversarial input perturbations. The error statistics are shown in Table~\ref{tab:poisson} below.

\begin{table}[H]
	\centering
	\caption{Relative $L^2$ error for the one-dimensional Poisson equation.
		The clean row uses $\Sop(f)$ as the reference, whereas the perturbed rows use $\Sop(f+\delta_{\mathrm{com}})$.
		Bold marks the smallest entry in each row.}
	\label{tab:poisson}
	\small
	\begin{tabular}{lrrr}
		\toprule
		Setting & PI-DeepONet & AdvPI & StablePDENet \\
		\midrule
		Clean
		& 0.066040 & 0.025944 & \textbf{0.014940} \\
		$\varepsilon=0.05$
		& 7.907976 & 0.040917 & \textbf{0.026344} \\
		$\varepsilon=0.10$
		& 15.596077 & 0.079383 & \textbf{0.045251} \\
		\bottomrule
	\end{tabular}
\end{table}

\subsection{\texorpdfstring{Burgers Equation}{Burgers Equation}}

We next consider the low-viscosity Burgers equation as a nonlinear
time-dependent benchmark,
\begin{equation}
	u_t+uu_x-\nu u_{xx}=0,
	\qquad
	(x,t)\in(0,1)\times(0,1],
\end{equation}
subject to the periodic boundary condition
\begin{equation}
	u(0,t)=u(1,t),
\end{equation}
and the initial condition
\begin{equation}
	u(x,0)=u_0(x).
\end{equation}
Here, $\nu=0.01$ denotes the viscosity coefficient. In the low-viscosity regime,
the solution may develop steep transition layers and shock-like structures,
making its long-time behavior sensitive to perturbations in the initial
condition. This example therefore provides a nonlinear and dynamical test for
evaluating whether the proposed method can preserve the intrinsic
perturbation response of the underlying PDE while suppressing additional
model-induced instability.
The neural operator is trained to approximate the solution mapping
\begin{equation}
	\mathcal{S}:u_0\mapsto u(\cdot,\cdot).
\end{equation}
In this experiment, the initial condition $u_0(x)$ is generated from a Gaussian random field with distribution
\begin{equation*}
	u_0\sim
	\mathcal{N}\left(
	0,25^2\left(-\Delta+5^2I\right)^{-4}
	\right).
\end{equation*}
We use $1000$ training functions, a function batch size of $50$, and $1000$ points for the PDE residual. The comparison set
contains $1000$ functions, and each output is evaluated on $101$ uniformly
spaced points. Burgers perturbations satisfy periodic endpoint compatibility. The reference trajectory is generated on an internal spatial grid of 4096 points before sampling.

%To provide a physically interpretable measure of perturbation propagation,
%we also monitor the displacement of the steepest solution front. At a given
%time $t$, the front location is defined as
%\begin{equation}
%	x_{\mathrm{s}}(t)
%	=
%	\arg\max_{x}
%	\left|
%	\partial_x u(x,t)
%	\right|.
%\end{equation}
%The perturbation-induced displacement is
%\begin{equation}
%	\Delta x_{\mathrm{s}}(t)
%	=
%	x_{\mathrm{s}}^{\delta}(t)
%	-
%	x_{\mathrm{s}}(t),
%\end{equation}
%and the shock-displacement error of the neural operator is measured by
%\begin{equation}
%	E_{\mathrm{shock}}(t)
%	=
%	\left|
%	\Delta x_{\mathrm{s}}^{\mathrm{NN}}(t)
%	-
%	\Delta x_{\mathrm{s}}^{\mathrm{ref}}(t)
%	\right|.
%\end{equation}
%
%We perform experiments for several viscosity coefficients in the
%low-viscosity regime, with particular attention to the growth of
%$\kappa_{\mathcal{E}}$ and $E_{\mathrm{shock}}$ as the viscosity decreases.
%The reference solutions are generated using a high-resolution
%Fourier pseudo-spectral solver with periodic boundary conditions. This
%experiment is designed to test whether StablePDENet can accurately reproduce
%the nonlinear propagation of small initial-condition perturbations,
%especially when such perturbations lead to appreciable displacement of
%shock-like structures.

\begin{figure}[htb]
	\centering
	\includegraphics[width=\linewidth]{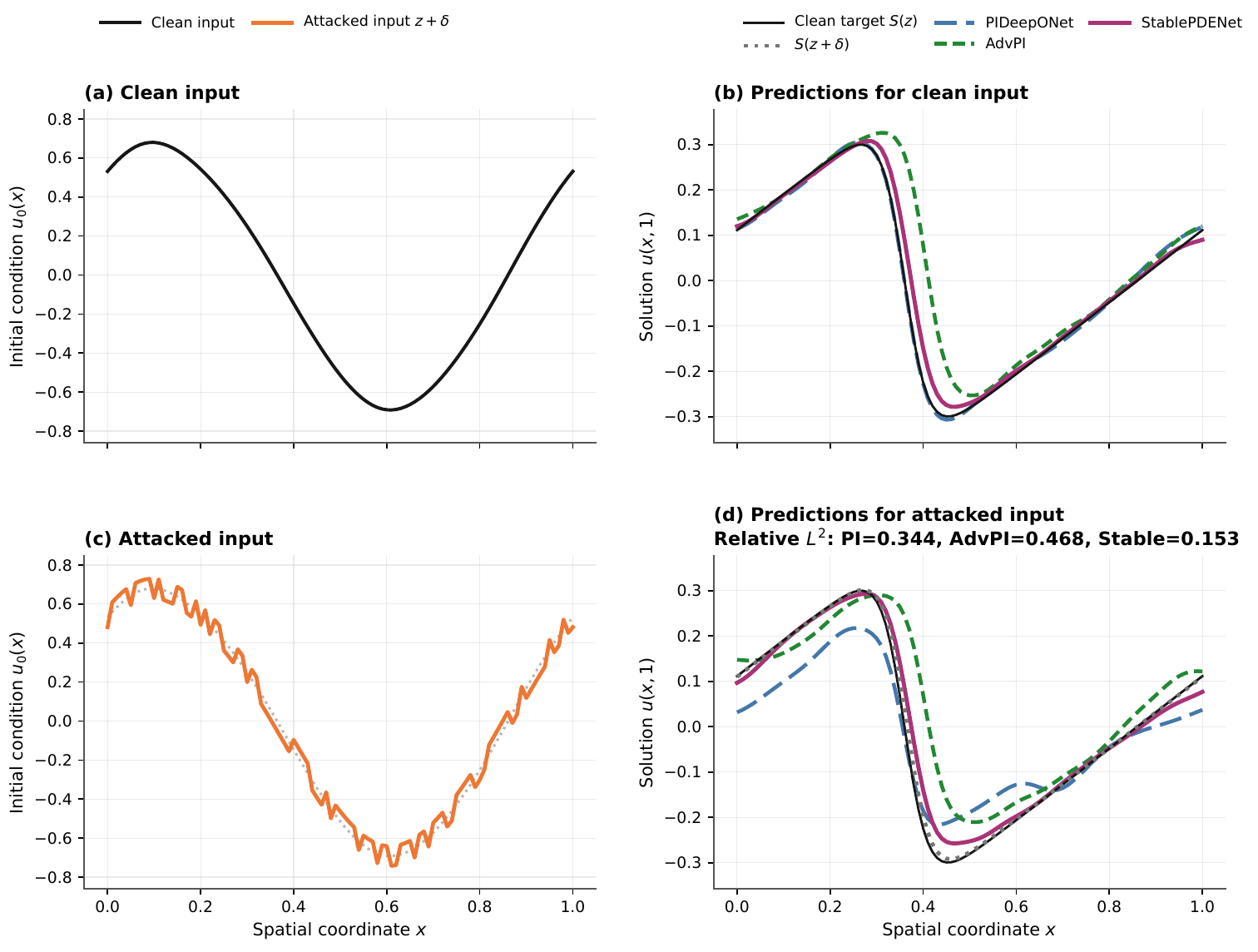}
	\caption{Burgers equation with initial condition perturbations. Left: original and adversarially perturbed initial conditions; right: predictions and reference solutions for the original and perturbed initial conditions. The perturbed initial condition is shared unchanged across the three models.}
	\label{fig:burgers}
\end{figure}

\begin{figure}[tb]
	\centering
	\includegraphics[width=\linewidth]{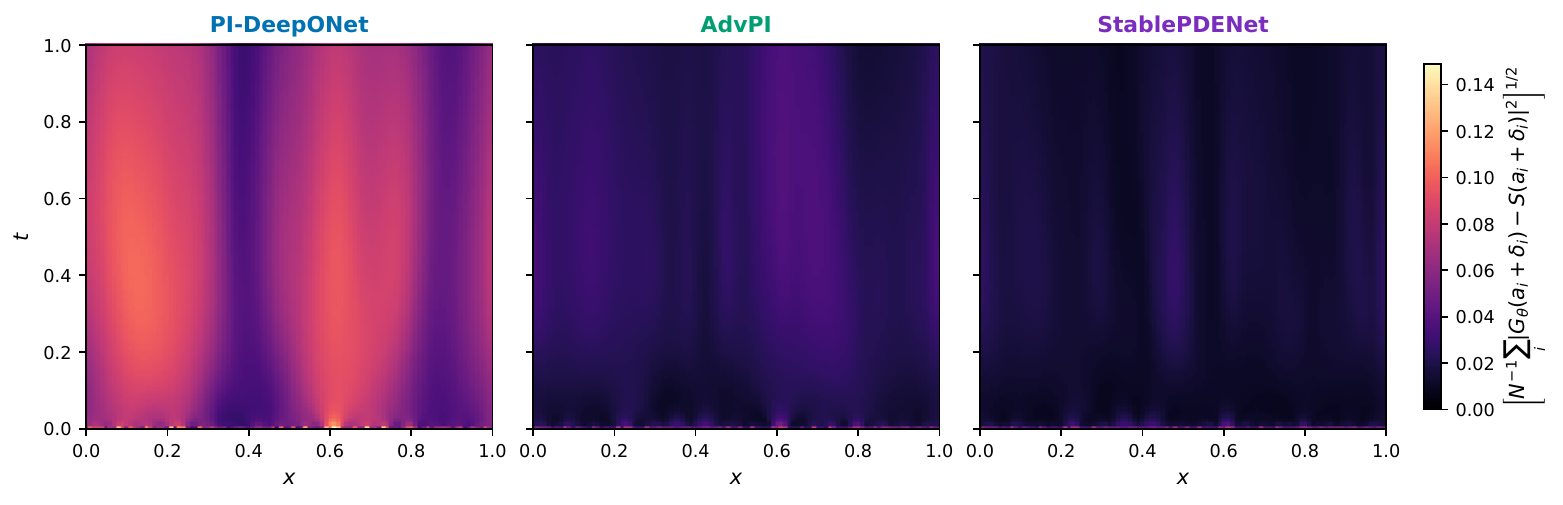}
	\caption{Space--time solution error for the Burgers equation with perturbed initial condition.}
	\label{fig:burgers_spacetime}
\end{figure}

Figures~\ref{fig:burgers} and~\ref{fig:burgers_spacetime} show the results for a representative test sample. The left column shows the original initial condition $u_0(x)$ and its adversarially perturbed counterpart $u_0(x)+\delta(x)$. The right column compares the three model predictions with the reference solutions. Our method maintains high accuracy under adversarial input perturbations. The error statistics are shown in Table~\ref{tab:burgers} below.

\begin{table}[H]
	\centering
	\caption{Full space--time relative $L^2$ error for the viscous Burgers equation.
		The clean row uses $\Sop(u_0)$ as the reference, whereas the perturbed rows use $\Sop(u_0+\delta_{\mathrm{com}})$.
		Bold marks the smallest entry in each row.}
	\label{tab:burgers}
	\small
	\begin{tabular}{lrrr}
		\toprule
		Setting & PI-DeepONet & AdvPI & StablePDENet \\
		\midrule
		Clean
		& \textbf{0.063751} & 0.072669 & 0.072235 \\
		$\varepsilon=0.05$
		& 0.434480 & 0.151233 & \textbf{0.095975} \\
		$\varepsilon=0.10$
		& 0.892059 & 0.283461 & \textbf{0.156127} \\
		\bottomrule
	\end{tabular}
\end{table}

\subsection{Helmholtz Equation with Coefficient-space Perturbations}
\label{numsection:helmholtz}

\begingroup
\setlength{\emergencystretch}{3pt}
We next consider the one-dimensional Helmholtz equation with homogeneous
Dirichlet boundary conditions,\par
\endgroup
\begin{equation}
	\begin{cases}
		-u_{xx}(x)-\kappa^2u(x)=f(x),
		&x\in(0,1),\\
		u(0)=u(1)=0.
	\end{cases}
	\label{eq:helmholtz}
\end{equation}
where $\kappa$ is the wavenumber. The objective
is to learn the solution operator
\begin{equation}
	\Sop_{\kappa}:f\longmapsto u,
	\qquad
	x\in[0,1].
\end{equation}
Unlike the time-dependent examples, the forcing is represented directly in
the orthonormal Fourier sine basis
\begin{equation}
	\phi_n(x)=\sqrt{2}\sin(n\pi x),
	\qquad
	f_{\bm c}(x)=\sum_{n=1}^{K}c_n\phi_n(x),
\end{equation}
and we set $K=10$. For a non-resonant value of $\kappa$, the reference solution is available in
closed form,
\begin{equation}
	\Sop_{\kappa}(f_{\bm c})(x)
	=
	\sum_{n=1}^{K}
	\frac{c_n}{(n\pi)^2-\kappa^2}\phi_n(x).
	\label{eq:helmholtz_exact_sine}
\end{equation}

%For coefficients supported on the first $K$ orthonormal sine modes, the exact Lipchitz constants are
%\begin{align}
%	C_{K}(\kappa)
%	&=\max_{1\le n\le K}
%	\left|(n\pi)^2-\kappa^2\right|^{-1},
%	\label{eq:helm-l2-constant}
%\end{align}
%In particular, for $\kappa=3.5\pi$ the spectral gap is
%\begin{equation}
%	\min_{1\leq n\leq K}
%	\left|(n\pi)^2-\kappa^2\right|
%	=3.25\pi^2\approx 32.076,
%\end{equation}
%so none of the retained modes is resonant. And the appximal Liphchitz constant is
%\begin{equation}
%	\frac{
	%		\|\mathcal G_\theta(f+\alpha\phi_{n^*})
	%		-\mathcal G_\theta(f)\|_{L^2}
	%	}{
	%		|\alpha|
	%	}.
%\end{equation}

The sine coefficients form the input to the branch network. Their
standard deviations decay with frequency according to
\begin{equation}
	\sigma_n
	=
	\frac{\left(1+(n/5)^2\right)^{-1}}
	{\left[
		\sum_{j=1}^{10}
		\left(1+(j/5)^2\right)^{-2}
		\right]^{1/2}},
	\qquad
	c_n\sim\mathcal{N}(0,\sigma_n^2).
\end{equation}
The training and validation functions contain modes $1$--$10$. We use $1000$ training functions,
$200$ validation functions, a function batch size of $50$, and $1000$
one-dimensional Hammersley points for the PDE residual. The comparison set
contains $2000$ functions, and each output is evaluated on $257$ uniformly
spaced points.

For the response analysis below, a coefficient perturbation
$\delta\bm c$ supported on the first $K_{\mathrm{adv}}$ sine modes induces
\begin{equation}
	\delta f(x)
	=
	\sum_{n=1}^{K_{\mathrm{adv}}}
	\delta c_n\phi_n(x),
	\qquad
	f_{\bm c+\delta\bm c}(x)
	=
	f_{\bm c}(x)+\delta f(x).
\end{equation}
Because the sine basis is orthonormal, Parseval's identity yields
\begin{equation}
	\|\delta f\|_{L^2(0,1)}^2
	=
	\sum_{n=1}^{K_{\mathrm{adv}}}|\delta c_n|^2
	=
	\|\delta\bm c\|_2^2
	=
	K_{\mathrm{adv}}
	\|\delta\bm c\|_{2,K_{\mathrm{adv}}}^2.
	\label{eq:helmholtz-coefficient-norm}
\end{equation}
Thus, before division by $K_{\mathrm{adv}}$, the Euclidean norm of the
coefficient vector is identical to the physical $L^2(0,1)$ norm of the
induced forcing perturbation.
By linearity of \eqref{eq:helmholtz}, the corresponding exact response is
\begin{equation}
	\begin{aligned}
		\Sop_{\kappa}(f_{\bm c}+\delta f)
		-\Sop_{\kappa}(f_{\bm c})
		&=
		\Sop_{\kappa}(\delta f)\\
		&=
		\sum_{n=1}^{K_{\mathrm{adv}}}
		\frac{\delta c_n}{(n\pi)^2-\kappa^2}\phi_n(x).
	\end{aligned}
	\label{eq:exact-response}
\end{equation}

\paragraph{Different wavenumber settings.}
\label{numsection:helmholtz_nearresonant_part}

We next keep the architecture, data representation, collocation rule,
training schedule, and adversarial budgets unchanged, but the wavenumber takes three different values,
\begin{equation}
	\kappa=3\pi+\eta,
	\qquad
	\eta\in\{0.5\pi,0.5,0.1\}.
	\label{eq:helmholtz_nearresonant_kappa}
\end{equation}
These cases range from non-resonant to near-resonant regimes rather than exactly resonant because
$\eta>0$. At $\eta=0$, the inverse operator is singular on the third sine
mode unless the corresponding forcing coefficient vanishes.

We set $K=K_{\mathrm{adv}}=10$, so the admissible coefficient subspace
contains the sensitive third mode while allowing perturbations to combine all
ten retained modes. Training maximizes the physics loss in
\eqref{eq:training-attack} with $\varepsilon=0.01$, whereas evaluation uses the
common-perturbation objective \eqref{eq:test-attack} with
$\varepsilon\in\{0.05,0.10\}$.
The boundary residual remains softly enforced using the same
resonance-scaled rule. Since
$\sin^2(3\pi+\eta)=\sin^2(\eta)$, its weight is
\begin{equation}
	\lambda_{\mathrm{BC}}(\eta)
	=
	\frac{2}
	{\max\{\sin^2(\eta),10^{-8}\}}.
	\label{eq:helmholtz_nearresonant_bc_weight}
\end{equation}
The inverse-square factor maintains a comparable boundary penalty near resonance, while the
floor $10^{-8}$ prevents numerical overflow.
This scaling prevents the boundary block from becoming negligible relative
to the amplified PDE response near resonance.

\begin{figure}[htb]
	\centering
	\includegraphics[width=\linewidth]{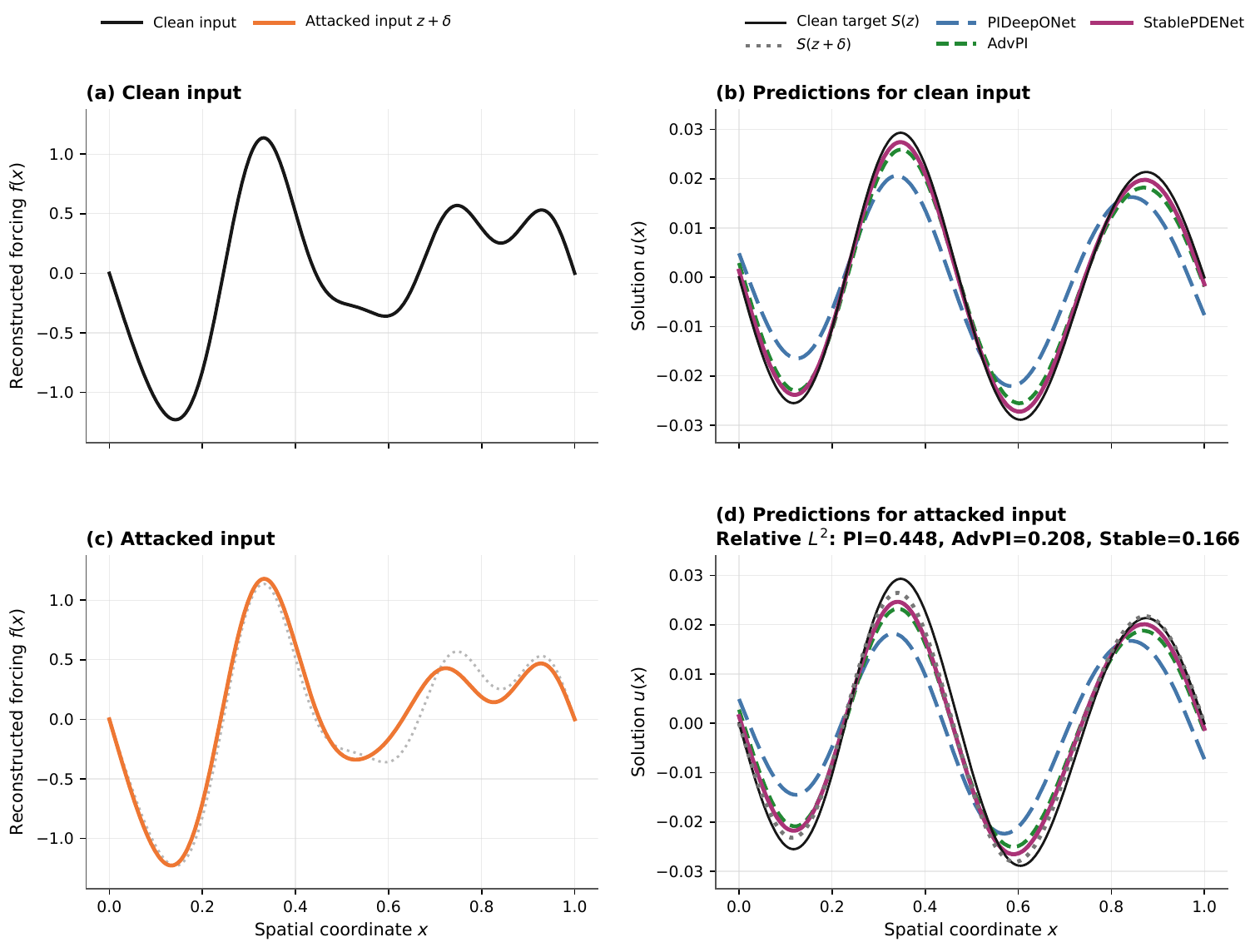}
	\caption{Helmholtz equation with coefficient-space perturbations. Left: original and adversarially perturbed source terms; right: predictions and reference solutions for the original and perturbed sources. All models are evaluated using the same perturbed coefficient vector.}
	\label{fig:helm_diffkappa}
\end{figure}

Table~\ref{tab:helmholtz_id10_perturbed_reference} reports the test results for the Helmholtz equation across different $\kappa$.
For each test input, the common coefficient perturbation is generated
once using the PI-DeepONet objective in Eq.~\eqref{eq:test-attack} and is supplied
unchanged to PI-DeepONet, AdvPI, and StablePDENet. The three predictions are
therefore compared under an identical coefficient perturbation and against the
same recomputed physical reference. The StablePDENet gives the smallest pooled error in
all nine rows. For all cases, it improves both clean
accuracy and attacked accuracy over AdvPI. Taken together, the
non-resonant experiment primarily tests adversarial sensitivity, while the near-resonant experiment additionally probes the
intrinsic conditioning of the physical solution operator.
Figure~\ref{fig:helm_diffkappa} shows the results for a representative sample at $\kappa=3.5\pi$. Our method maintains high accuracy on both clean and adversarially perturbed inputs.

\begin{table}[H]
	\centering
	\caption{Relative $L^2$ error with respect to the clean and perturbed reference. Clean rows use the reference
		$\mathcal{S}_\kappa(f_{\boldsymbol{c}})$, whereas attacked rows use the
		recomputed physical reference
		$\mathcal{S}_\kappa(f_{\bm c+\delta\bm c})$. Bold marks the smallest entry in each row.}
	\label{tab:helmholtz_id10_perturbed_reference}
	\small
	\begin{tabular}{llrrr}
		\toprule
		$\kappa$ & Setting & PI-DeepONet & AdvPI & StablePDENet \\
		\midrule
		\multirow{3}{*}{$3.5\pi$}
		& Clean & 0.326010 & 0.089857 & \textbf{0.064999} \\
		& $\varepsilon=0.05$ & 0.346506 & 0.086149 & \textbf{0.061379} \\
		& $\varepsilon=0.10$ & 0.367778 & 0.084560 & \textbf{0.061122} \\
		\midrule
		\multirow{3}{*}{$3\pi+0.5$}
		& Clean & 0.142330 & 0.068950 & \textbf{0.041629} \\
		& $\varepsilon=0.05$ & 0.151115 & 0.069138 & \textbf{0.041643} \\
		& $\varepsilon=0.10$ & 0.160183 & 0.069242 & \textbf{0.041646} \\
		\midrule
		\multirow{3}{*}{$3\pi+0.1$}
		& Clean & 0.177143 & 0.042738 & \textbf{0.027478} \\
		& $\varepsilon=0.05$ & 0.182066 & 0.043542 & \textbf{0.028372} \\
		& $\varepsilon=0.10$ & 0.187390 & 0.044627 & \textbf{0.029578} \\
		\bottomrule
	\end{tabular}
\end{table}

\paragraph{Comparison of Lipschitz constants.}

For coefficients supported on the first $K$ orthonormal sine modes, the exact Lipschitz constants are
\begin{align}
	C_K(\kappa)
	&=
	\max_{1\leq n\leq K}
	\left|(n\pi)^2-\kappa^2\right|^{-1}.
	\label{eq:helm-l2-constant}
\end{align}
For the third mode, the denominator in
Eq.~\eqref{eq:helmholtz_exact_sine} becomes
\begin{equation}
	(3\pi)^2-(3\pi+\eta)^2
	=
	-(6\pi\eta+\eta^2).
\end{equation}
Consequently, the relevant spectral gap and exact modal gain are
\begin{equation}
	g_\eta=6\pi\eta+\eta^2,
	\qquad
	\left|(3\pi)^2-\kappa^2\right|^{-1}
	=
	g_\eta^{-1}.
	\label{eq:helmholtz_nearresonant_gain}
\end{equation}
Accordingly, the same coefficient perturbation
can produce an increasingly large solution response as $\eta$ decreases.
Theorem~\ref{the:linear} makes explicit both the mechanism and the
difficulty of the near-resonant regime: the residual-sensitivity loss controls the
response-consistency error, but with an amplification factor $1/g_\eta$ that
diverges as $\kappa^2$ approaches the spectrum.

For a neural operator evaluated at $257$ points, we form the sample Jacobian $J_\theta(\bm c)$ and apply trapezoidal weights $W$. The local Lipschitz constant is $\sigma_{\max}(W^{1/2}J_\theta)$.
Table~\ref{tab:helm-lipschitz} shows that the StablePDENet is closest to the exact constant for all three settings as $\kappa$ decreases. This indicates that our method learns an accurate response of the Helmholtz system.

\begin{table}[H]
	\centering
	\caption{Exact and neural local Lipschitz constants.}
	\label{tab:helm-lipschitz}
	\small
	\begin{tabular}{lrrrr}
		\toprule
		$\kappa$ & Exact $C_K(\kappa)$ & PI-DeepONet & AdvPI & StablePDENet \\
		\midrule
		$3.5\pi$ & 0.031176 & 0.02706 & 0.02965 & \textbf{0.02980} \\
		$3\pi+0.5$ & 0.103362 & 0.08863 & 0.09614 & \textbf{0.09915} \\
		$3\pi+0.1$ & 0.527717 & 0.43538 & 0.50643 & \textbf{0.51405} \\
		%		$3\pi+0.02$ & 2.649771 & 0.65664 & 1.10512 & \textbf{1.21327} \\
		\bottomrule
	\end{tabular}
\end{table}

\subsection{Physical versus model-induced response}

All response gains are evaluated using the same common perturbation
$\delta_{\mathrm{com}}$ for every
$m\in\mathcal{M}_{\mathrm{eval}}$. The physical response gain is
\begin{equation}
	\kappa_{\mathcal S}
	=
	\frac{
		\left\|
		\mathcal S(a+\delta_{\mathrm{com}})
		-
		\mathcal S(a)
		\right\|
	}{
		\|\delta_{\mathrm{com}}\|_{L^2}
	},
\end{equation}
which measures the intrinsic sensitivity of the exact solution operator.
The learned-model response gain is
\begin{equation}
	\kappa_{\mathcal G,m}
	=
	\frac{
		\left\|
		\mathcal G_m(a+\delta_{\mathrm{com}})
		-
		\mathcal G_m(a)
		\right\|
	}{
		\|\delta_{\mathrm{com}}\|_{L^2}
	},
\end{equation}
which measures the response magnitude predicted by model $m$.
The response-consistency error is
\begin{equation}
	\kappa_{\mathcal E,m}
	=
	\frac{
		\left\|
		\left[
		\mathcal G_m(a+\delta_{\mathrm{com}})
		-
		\mathcal G_m(a)
		\right]
		-
		\left[
		\mathcal S(a+\delta_{\mathrm{com}})
		-
		\mathcal S(a)
		\right]
		\right\|
	}{
		\|\delta_{\mathrm{com}}\|_{L^2}
	},
\end{equation}
which measures the discrepancy between the learned and exact perturbation responses.
The residual-response gain is
\begin{equation}
	\kappa_{r,m}
	=
	\frac{
		\left[
		\displaystyle
		\sum_{q\in\mathcal B}
		w_q
		\left\|
		\mathbf r_{m,q}(a+\delta_{\mathrm{com}})
		-
		\mathbf r_{m,q}(a)
		\right\|_{2,N_q}^2
		\right]^{1/2}
	}{
		\|\delta_{\mathrm{com}}\|_{L^2}
	}.
\end{equation}
Here, $\mathbf r_{m,q}$ denotes the sampled residual vector of model $m$ in block $q$.
Its block-weighted squared form is controlled by
$\mathcal Q_{\mathcal B}$.
For a fixed test sample, $\kappa_{\mathcal S}$ is identical across all
models, whereas $\kappa_{\mathcal G,m}$,
$\kappa_{\mathcal E,m}$, and $\kappa_{r,m}$ are model dependent.
Stability is indicated primarily by a small
$\kappa_{\mathcal E,m}$ together with
$\kappa_{\mathcal G,m}\approx\kappa_{\mathcal S}$, rather than by a small
$\kappa_{\mathcal G,m}$.

\begin{table}[htbp]
	\centering
	\caption{Perturbation-response gains for the four equations (Helmholtz with $\kappa=3.5\pi$). Within each equation and perturbation radius, all models are evaluated under the same common perturbation.
		Bold indicates the smallest $\kappa_{\mathcal E,m}$ and $\kappa_{r,m}$
		for each equation and perturbation radius.}
	\label{tab:pooled_response_gains}
	\small
	\begin{tabular}{lllrrrr}
		\toprule
		Equation & $\varepsilon$ & Model
		& $\kappa_{\mathcal S}$
		& $\kappa_{\mathcal G,m}$
		& $\kappa_{\mathcal E,m}$
		& $\kappa_{r,m}$ \\
		\midrule
		
		\multirow{6}{*}{Heat}
		& \multirow{3}{*}{$0.05$}
		& PI-DeepONet
		& \multirow{3}{*}{0.112741}
		& 9.644770 & 9.632763 & 10.151563 \\
		& & AdvPI
		& & 0.828631 & 0.822415 & 1.385824 \\
		& & StablePDENet
		& & 0.302234 & \textbf{0.319088} & \textbf{1.063685} \\
		\cmidrule(lr){2-7}
		& \multirow{3}{*}{$0.10$}
		& PI-DeepONet
		& \multirow{3}{*}{0.112936}
		& 9.567076 & 9.559970 & 10.074741 \\
		& & AdvPI
		& & 0.818964 & 0.816574 & 1.377524 \\
		& & StablePDENet
		& & 0.357539 & \textbf{0.374206} & \textbf{1.087987} \\
		
		\midrule
		\multirow{6}{*}{Poisson}
		& \multirow{3}{*}{$0.05$}
		& PI-DeepONet
		& \multirow{3}{*}{0.008403}
		& 9.719419 & 9.712301 & 12.814424 \\
		& & AdvPI
		& & 0.048053 & 0.052287 & 0.924988 \\
		& & StablePDENet
		& & 0.028877 & \textbf{0.026447} & \textbf{0.854243} \\
		\cmidrule(lr){2-7}
		& \multirow{3}{*}{$0.10$}
		& PI-DeepONet
		& \multirow{3}{*}{0.007598}
		& 9.649538 & 9.643112 & 12.715485 \\
		& & AdvPI
		& & 0.048985 & 0.053128 & 0.920936 \\
		& & StablePDENet
		& & 0.025587 & \textbf{0.024742} & \textbf{0.869966} \\
		
		\midrule
		\multirow{6}{*}{Burgers}
		& \multirow{3}{*}{$0.05$}
		& PI-DeepONet
		& \multirow{3}{*}{0.328323}
		& 1.635470 & 1.376695 & 7.169455 \\
		& & AdvPI
		& & 0.654247 & 0.421779 & 4.434917 \\
		& & StablePDENet
		& & 0.422333 & \textbf{0.207195} & \textbf{4.208414} \\
		\cmidrule(lr){2-7}
		& \multirow{3}{*}{$0.10$}
		& PI-DeepONet
		& \multirow{3}{*}{0.314095}
		& 1.690015 & 1.454408 & 7.573011 \\
		& & AdvPI
		& & 0.658344 & 0.445795 & 4.505868 \\
		& & StablePDENet
		& & 0.426380 & \textbf{0.225384} & \textbf{4.260864} \\
		
		\midrule
		\multirow{6}{*}{Helmholtz}
		& \multirow{3}{*}{$0.05$}
		& PI-DeepONet
		& \multirow{3}{*}{0.014250}
		& 0.014014 & 0.003544 & 0.023837 \\
		& & AdvPI
		& & 0.012757 & 0.002064 & 0.011848 \\
		& & StablePDENet
		& & 0.012810 & \textbf{0.001830} & \textbf{0.008719} \\
		\cmidrule(lr){2-7}
		& \multirow{3}{*}{$0.10$}
		& PI-DeepONet
		& \multirow{3}{*}{0.014253}
		& 0.014100 & 0.003650 & 0.023990 \\
		& & AdvPI
		& & 0.012718 & 0.002108 & 0.012275 \\
		& & StablePDENet
		& & 0.012777 & \textbf{0.001868} & \textbf{0.008746} \\
		
		\bottomrule
	\end{tabular}
\end{table}

Table~\ref{tab:pooled_response_gains} reports
$\kappa_{\mathcal S}$ and the model-dependent gains
$\kappa_{\mathcal G,m}$, $\kappa_{\mathcal E,m}$, and
$\kappa_{r,m}$ for the four benchmark equations. The results show that
residual-sensitivity regularization suppresses spurious model responses while
preserving the intrinsic sensitivity of the physical solution operator. For
Heat and Poisson, $\kappa_{\mathcal E,m}>\kappa_{\mathcal S}$ for all
evaluated models, indicating that response consistency remains imperfect.
Closer agreement is obtained for Burgers and Helmholtz, particularly with the
StablePDENet. The Helmholtz results further show that
$\kappa_{\mathcal G,\PI}\approx\kappa_{\mathcal S}$ alone does not ensure
response consistency: PI-DeepONet has a larger
$\kappa_{\mathcal E,\PI}$, indicating a mismatch in response direction.
Overall, $\kappa_{\mathcal E,m}$ is the primary measure of response
consistency, while $\mathcal Q_{\mathcal B}$ controls a block-weighted
mean-squared counterpart of $\kappa_{r,m}^2$.

\section{Conclusion}\label{section:04}
In this work, we present a stable framework named StablePDENet for learning solution operators of differential equations that maintain accuracy while improving stability against input perturbations. The key idea lies in our adversarial training paradigm, which systematically combines a physics-based projected-gradient adversary with a normalized residual-sensitivity regularizer. 
Instead of straightforwardly minimizing the Lipschit constant of the neural operator,  the StablePDENet approximates the exact local Lipschitz constant of the PDE operator.
The analysis uses the error operator to relate residual sensitivity to the
Lipschitz response of the learned-solution error. The estimate follows from
bounded invertibility in the linear case and holds locally under
inverse-stability and neighborhood assumptions in the nonlinear case; it
therefore provides conditional theoretical support rather than a global
post-training stability certificate.
The effectiveness of our method has been evaluated through several experiments, showing that our method not only preserves the accuracy of the baseline methods on unperturbed inputs, but also demonstrates improved stability to adversarial
input perturbations. The numerical results also demonstrate that the StablePDENet can effectively improve the generalization accuracy for operator learning.
The Helmholtz experiments are particularly
informative: StablePDENet achieves the lowest prediction errors and provides
the closest empirical approximation to the exact local Lipschitz constant of
the solution operator.
 The demonstrated improvements in stability suggest promising potential for deploying neural operators in real-world applications where input uncertainties and measurement noise are inevitable.

\setcounter{table}{0}   %从0开始编号，显示出来表会A1开始编号
\setcounter{figure}{0}
\setcounter{algorithm}{0}
%定义编号格式，在数字序号前加字符“A"
\renewcommand{\thetable}{A\arabic{table}}
\renewcommand{\thefigure}{A\arabic{figure}}
\renewcommand{\thealgorithm}{A\arabic{algorithm}}
\renewcommand{\theHalgorithm}{A\arabic{algorithm}}

\section*{Acknowledgement}

The work was supported in part by the National Key Research and Development Program of China (No. 2025YFA1016800) and the Project of Hetao
Shenzhen-HKUST Innovation Cooperation Zone HZQB-KCZYB-2020083. 

\bibliographystyle{unsrt}
\bibliography{sample}

\end{document}